\documentclass[pdflatex,sn-mathphys-num]{sn-jnl}

\usepackage{graphicx,subfig}
\usepackage{amsmath,amssymb,amsfonts,amsthm}%
\usepackage{mathrsfs}%
\usepackage[title]{appendix}%
\usepackage{xcolor}%
\usepackage{textcomp}%
\usepackage{manyfoot}%
\usepackage{booktabs}%
\usepackage{algorithm,algorithmicx,algpseudocode}%
\usepackage{listings}%
\usepackage{multirow,longtable}
\usepackage{bm,textgreek}

\newtheoremstyle{thmstyleone}
  {3pt}
  {3pt}
  {\itshape}
  {}
  {\bfseries}
  {.}
  { }
  {}

\theoremstyle{thmstyleone}
\newtheorem{theorem}{Theorem}

\title{PUAL: A Classifier on Trifurcate Positive-Unlabeled Data}

\author[1]{\fnm{Xiaoke} \sur{Wang}}\email{xiaoke.wang.18@ucl.ac.uk}

\author[2]{\fnm{Xiaochen} \sur{Yang}}\email{xiaochen.yang@glasgow.ac.uk}

\author*[3]{\fnm{Rui} \sur{Zhu}}\email{rui.zhu@city.ac.uk}

\author[1]{\fnm{Jing-Hao} \sur{Xue}}\email{jinghao.xue@ucl.ac.uk}

\affil[1]{\orgdiv{Department of Statistical Science}, \orgname{University College London}, \orgaddress{\street{Gower Street}, \city{London}, \postcode{WC1E 6BT}, \state{England}, \country{UK}}}

\affil[2]{\orgdiv{School of Mathematics \& Statistics}, \orgname{University of Glasgow},
\orgaddress{\street{University Place}, \city{Glasgow}, \postcode{G12 8QQ}, \state{Scotland}, \country{UK}}}

\affil*[3]{\orgdiv{Bayes Business School}, \orgname{City, University of London}, 
\orgaddress{\street{106 Bunhill Row}, \city{London}, \postcode{EC1Y 8TZ}, \state{England}, \country{UK}}}

\def\*#1{\bm{#1}}

\begin{document}
\maketitle

\begin{abstract} 
 \ \textbf{Abstract}: Positive-unlabeled (PU) learning aims to train a classifier using the data containing only labeled-positive instances and unlabeled instances. However, existing PU learning methods are generally hard to achieve satisfactory performance on trifurcate data, where the positive instances distribute on both sides of the negative instances. To address this issue, firstly we propose a PU classifier with asymmetric loss (PUAL), by introducing a structure of asymmetric loss on positive instances into the objective function of the global and local learning classifier. Then we develop a kernel-based algorithm to enable PUAL to obtain non-linear decision boundary. We show that, through experiments on both simulated and real-world datasets, PUAL can achieve satisfactory classification on trifurcate data.
\end{abstract}

\section{Introduction}\label{introduction}

PU learning is to train a classifier from PU data, which only contain labeled-positive instances and unlabeled instances, i.e., the PU data lack labeled-negative instances for training. Recently, there are more and more PU data occurring in practice, such as deceptive review detection~\cite{ren2014positive}, text categorization~\cite{li2014spotting} and remote sensing classification~\cite{li2010positive, dai2023gradpu}. 

The main difficulty in PU learning is the lack of labeled-negative instances.  A natural way to deal with this issue is to pick the instances highly likely to be negative from the unlabeled set and treat them as negative; then the obtained dataset will contain labeled-positive, labeled-negative and unlabeled instances; and thus finally PU learning can be converted to classical semi-supervised learning. The methods following this idea are often  referred to  as two-step methods~\cite{liu2002partially,yu2004pebl,li2003learning,he2018instance,liu2021new,liu2022new,ienco2012context,liu2014clustering}, 
which can also be generalized to have more steps of further iterative training~\cite{chaudhari2012learning,ke2012building,he2023novel,xu2022split, dorigatti2022positive}.

However, the accuracy of such a multi-step method relies heavily on the accuracy of the algorithm applied in the first step to pick reliable negative instances~\cite{liang2023positive}. This drawback motivated studies to train a PU classifier in a single step, termed one-step methods, which can be further categorized into inconsistent PU learning methods and  consistent PU learning methods, depending on if the objective function is a consistent estimator of an expected loss to classify an unknown instance frorm the population.

A pioneer consistent PU learning method is the unbiased PU learning (uPU)~\cite{du2014analysis}. Subsequently, the non-negative PU learning (nnPU)~\cite{kiryo2017positive} was proposed by taking the absolute value of the estimated average loss on the negative set in the objective function of uPU  for better convergence of the classifier training.  Furthermore, imbalanced nnPU (imbalancednnPU)~\cite{su2021positive} was proposed to address imbalanced PU training set.
Nevertheless, the training of the consistent methods often required more information out of the PU dataset, e.g., the class prior or the distribution of population. 

Early attempt of  the inconsistent PU learning methods was the biased support vector machine (BSVM)~\cite{liu2003building} based on the classic supervised support vector machine (SVM)~\cite{cortes1995support}, assigning a high weight to the average loss of the labeled-positive instances and a low weight to the average loss of the unlabeled instances in the objective function. Subsequently, weighted unlabeled samples SVM (WUS-SVM)~\cite{liu2008partially} was proposed to assign a distinct weight  to each of the unlabeled instances according to the likelihood of this unlabeled instance to be negative. Then, the biased least squares SVM (BLS-SVM) was proposed in~\cite{ke2018biased}. It replaces the hinge loss with the squared loss in the objective of BSVM in case that too much importance is given to the unlabeled-positive instances treated as negative during the training of classifier. Then the local constraint was introduced to the objective function of BLS-SVM in~\cite{ke2018global}, encouraging the instances to be classified into the same class as its neighborhoods and the proposed method is called the global and local learning classifier (GLLC). Moreover, the large-margin label-calibrated SVM (LLSVM)~\cite{gong2019large} was proposed to further alleviate the bias by introducing the hat loss to the objective function, where the hat loss measures the gap between the inverse tangent of the predictive score function (without intercept) and a certain threshold. 

\begin{figure}[htbp]    
    \centering           
    \includegraphics[width=\textwidth]{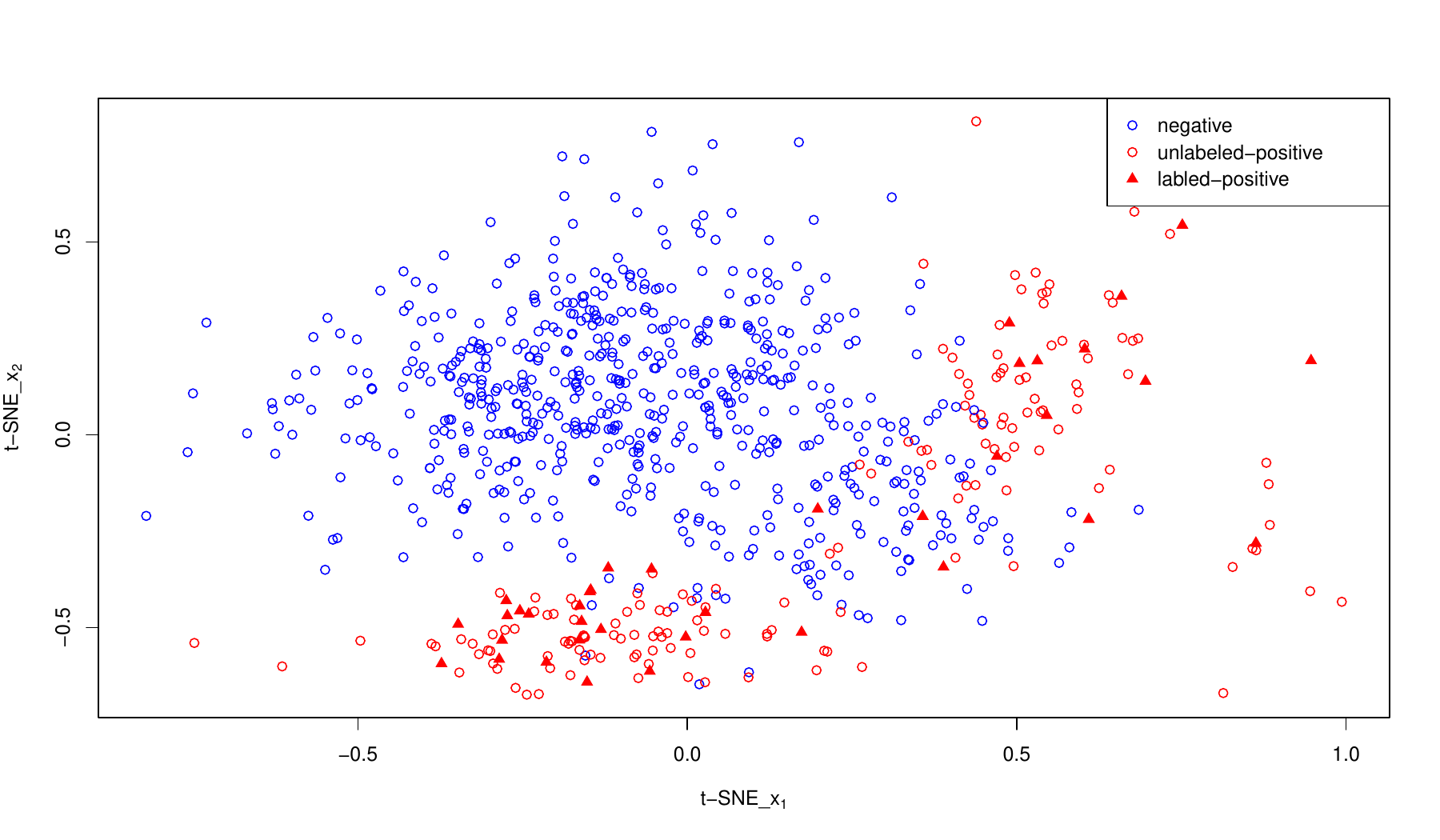}  
    \caption{ The 2-dimensional projection with t-SNE of dataset \textbf{wifi}, where the positive set is roughly constituted by two subsets distributing on both sides of negative instances; the perplexity for the training of t-SNE on \textbf{wifi} was set to 750.}          
    \label{eco}               
    \end{figure}

However, for trifurcate data where the positive set is roughly constituted by two subsets distributing on both sides of negative instances, existing PU learning methods are usually hard to achieve satisfactory performance. An example of the trifurcate PU datasets  is illustrated in Fig.~\ref{eco} by the 2-dimensional projection with the t-distributed stochastic neighbor embedding (t-SNE)~\cite{maaten2008visualizing} of dataset Wireless Indoor  Localization  (\textbf{wifi})~\cite{misc_wireless_indoor_localization_422}.

\begin{figure}[htbp]    
    \centering           
    \includegraphics[width=15.5cm, angle=0]{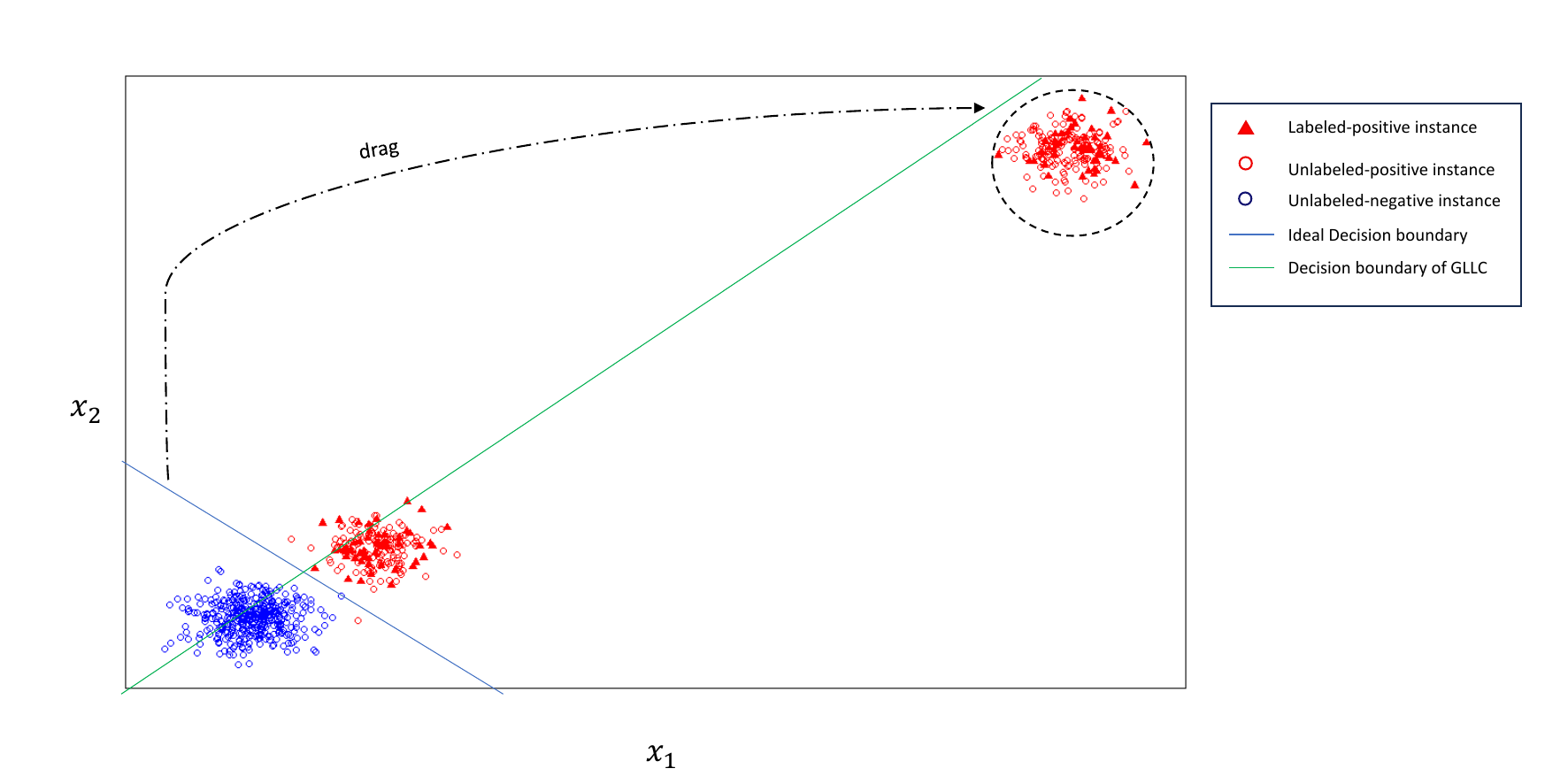}  
    \caption{A pattern of the linearly separable space constructed from the original trifurcate PU datasets via the kernel trick; $x_1$ and $x_2$ represent the mappings of the features. }          
    \label{demo_gllc}               
    \end{figure}

A classifier with non-linear decision boundary is needed for the classification of trifurcate datasets. However, when we apply kernel trick to obtain the non-linear decision boundary, the original trifurcate datasets may be converted to follow the pattern illustrated in Fig.~\ref{demo_gllc}, where the distances from the two positive subsets to the ideal decision boundary, as indicated by a solid blue line, are very different. In this case, using the squared loss (e.g. that used in GLLC) will impose quadratic penalty not only on the instances wrongly classified but also on all the instances correctly classified. Therefore, the labeled-positive instances correctly classified by, but far away from, the ideal decision boundary, as circled by the dashed lines in Fig.~\ref{demo_gllc},  can unfortunately generate large penalty via the squared loss and hence drag the ideal decision boundary towards the labeled-positive instances far away. This leads to the estimated decision boundary indicated in green solid line, which misclassifies many more instances than the ideal decision boundary. 

The aim of applying the squared loss in the objective function of GLLC was to ensure the importance given to the unlabeled-positive instances to be negative support vectors to be lower than the importance given to unlabeled-negative instances to be negative support vectors. Hence we note that it is not necessary to also use the  squared loss on the labeled-positive set. Furthermore, we note that, as the hinge loss does not penalise the instances lying on the correct side of the margin, if the hinge loss is applied to the labeled-positive set, the instances correctly classified but far away from the ideal decision boundary will not generate any wrong penalty. 

Therefore, motivated by the above analysis, we proposed a PU classifier with asymmetric loss (PUAL) on positive instances for better classification on the trifurcate PU datasets. The contribution of this paper can be summarized as follows. 

Firstly, in Section~\ref{motR}, we propose the methodology and algorithm  of PUAL, which can generate a linear decision boundary much closer to the ideal decision boundary for the datasets following the pattern in  Fig.~\ref{demo_gllc} but first in the original feature space. This is achieved by using the hinge loss for the labeled-positive instances and the squared loss for the unlabeled-instances (and thus the unlabeled-positive instances). 

Secondly, we develop a kernel-based algorithm to enable PUAL to obtain non-linear decision boundary in the original feature space, as detailed in Section~\ref{sec:kernel}. 

Thirdly, we conduct experiments on both simulated and real-world datasets to clearly verify the motivation and effectiveness of our proposed methods, as presented in Section~\ref{EXP}. 

\section{Related Work}
Firstly we provide some general definitions involved in the following related works. Suppose there are $n_p$ labeled-positive instances and $n_u$ unlabeled instances with $m$ features. Let feature matrix $\*{X}_{[ pu]}=(\*{x}_1,\dots,\*{x}_{n_p},\dots, \*{x}_{ n_p+n_u})^T\in\mathbb{R}^{(n_p+n_u)\times m}$, where the column vector $\*{x}_i\in\mathbb{R}^{m\times 1}$ denotes the vector of the features of the $i$th instance. Similarly, matrix $\*{X}_{[p]}=(\*{x}_1,\dots,\*{x}_{n_p})^T\in\mathbb{R}^{n_p\times m}$ denotes the feature matrix of the labeled-positive set while matrix $\*{X}_{[u]}=(\*{x}_{n_p+1},\dots,\*{x}_{n_p+n_u})^T\in\mathbb{R}^{n_u\times m}$ denotes the feature matrix of the unlabeled set.   The methods listed in this section are all aimed to train the following predictive score function for classification:
  \begin{equation}
  \label{prebsvm}
 f=\*{x}^T\*{\beta}+\beta_0,
  \end{equation}
where $\*{\beta}=(\beta_1,\beta_2,\dots,\beta_m)^T\in\mathbb{R}^{m\times1}$ and $\beta_0$ are the model parameters.


\subsection{BSVM}
In order to enable SVM to handle PU classification, BSVM~\cite{liu2003building} treats all the unlabeled instances as negative and assigns the loss of labeled-positive instances and the loss of unlabeled instances with different weights in the objective function. BSVM trains classifiers by solving the following objective function:
\begin{equation}\label{bsvm}
\begin{aligned}
\min\limits_{\*{\beta},\beta_0}&\frac{\lambda}{2}\*{\beta}^T\*{\beta} +C_p\*{1}_p^T{[\*{1}_p-(\*{X}_{[p]}\*{\beta}+\*{1}_p\beta_0)]}_+\\ 
&+C_u\*{1}_u^T{[\*{1}_u+(\*{X}_{[u]}\*{\beta}+\*{1}_u\beta_0)]}_+,
\end{aligned}
\end{equation}
where  $C_p$, $C_u$, and $\lambda$ are positive hyper-parameters, $[g(\cdot)]_+$ indicates the column vector of the maximum between each element of $g(\cdot)$ and 0, and $\*{1}_{p,u}=(\underbrace {1,1, \cdots ,1}_k)^T,k=n_p,n_u$.

\subsection{BLS-SVM}
One weakness of BSVM is that sometimes the hinge loss in the objective function of BSVM in Equation (\ref{bsvm}) selects  more unlabeled-positive instances than unlabeled-negative instances to be the support vectors for the negative class, which constructs  a decision boundary tending to misclassify the unlabeled-positive instances as negative. This is more likely to happen when there are many unlabeled-positive instances  close to the unlabeled negative instances. 

To deal with this issue, BLS-SVM was proposed by~\cite{ke2018biased} to force all training instances to contribute to the construction of the  decision boundary of the trained SVM by solving the following optimization problem: 
\begin{equation}
\label{blsvm}
\begin{aligned}
\min\limits_{\*{\beta},\beta_0}&\frac{\lambda}{2}\*{\beta}^T\*{\beta} +C_p{[\*{1}_p-(\*{X}_{[p]}\*{\beta}+\*{1}_p\beta_0)]}^T{[\*{1}_p-(\*{X}_{[p]}\*{\beta}+\*{1}_p\beta_0)]}\\ 
&+C_u{[\*{1}_u+(\*{X}_{[u]}\*{\beta}+\*{1}_u\beta_0)]}^T{[\*{1}_u+(\*{X}_{[u]}\*{\beta}+\*{1}_u\beta_0)]},
\end{aligned}
\end{equation}
where the squared loss replaces the hinge loss in BSVM on both labeled-positive set and unlabeled set. The objective function of BLS-SVM makes all the instances contribute to the construction of the decision boundary hence the importance given to the unlabeled-positive instances treated as negative is restricted~\cite{scott2009novelty}.


\subsection{GLLC}
The similarities between a training instance and its neighbors can also be treated as a factor for classification, the idea of which is named local learning~\cite{chapelle2009semi}. It is noted that the gap between PU learning and classical supervised learning on accuracy can be mitigated via GLLC~\cite{ke2018global}, which is a combination of BLS-SVM and local learning. The objective function of GLLC is given as
\begin{equation}
\label{objgllc}
 \begin{aligned}
\min\limits_{\*{\beta},\beta_0}&\frac{\lambda}{2}\*{\beta}^T\*{\beta} +C_p{[\*{1}_p-(\*{X}_{[p]}\*{\beta}+\*{1}_p\beta_0)]}^T{[\*{1}_p-(\*{X}_{[p]}\*{\beta}+\*{1}_p\beta_0)]}\\ 
&+C_u{[\*{1}_u+(\*{X}_{[u]}\*{\beta}+\*{1}_u\beta_0)]}^T{[\*{1}_u+(\*{X}_{[u]}\*{\beta}+\*{1}_u\beta_0)]}\\
&+(\*{X}_{[pu]}\*{\beta}+\*{1}_{pu}\beta_0)^T\*R(\*{X}_{[pu]}\*{\beta}+\*{1}_{pu}\beta_0),
\end{aligned}  
\end{equation}
where $\*R$ is the  similarity matrix for the instances and their neighbors, which can be calculated as follows. 

Firstly we need to calculate matrix $\*W$ by
\begin{equation}
\label{R1}
w_{i j}= \begin{cases}
\exp \left(-{\sigma}^{-1}(\*x_{i}-\*x_{j})^T (\*x_{i}-\*x_{j})\right) & \text{if the $i$th and $j$th instances are KNN of each other,}\\
0 & \text{otherwise,}
\end{cases}
\end{equation}
where $\sigma$ is a hyper-parameter to be selected.
Then defining $\*{1}_{pu}=(\underbrace {1,1, \cdots ,1}_k)^T,k=n_p+n_u$ and letting $\*w_{\cdot i}$ denote the $i$th column of matrix $\*W$  and $\*W^*$ denote a diagonal matrix with  the $i$th diagonal element equal to $\*1_{[pu]}^T\*w_{\cdot i}$, one can obtain
\begin{equation}
\label{RRR}
\*R=\frac{1}{(n_p+n_u)}(\*W^*-\*W).
\end{equation}


\section{Methodology}

\subsection{PUAL with Linear Decision Boundary}\label{motR}

\subsubsection{Objective Function}
 As discussed in Section \ref{introduction}, we propose to apply the hinge loss to the labeled-positive instances, so as to impose no penalty to the instances correctly classified but far away from the ideal decision boundary. Meanwhile, the squared loss needs to be applied to the unlabeled instances in case that too much importance is given to the unlabeled-positive instances treated as negative during the training of the classifier. Hence we propose a structure of asymmetric loss on positive instances to revise GLLC, so that the new method PUAL is able to address the issue of classification of trifurcate PU data. Moreover, a local constraint term of similarity is helpful for PU classification. Therefore, the unconstrained optimization problem of PUAL can be formulated as
    \begin{equation}\label{opt}
\begin{aligned}
\min\limits_{\*{\beta},\beta_0}&\frac{\lambda}{2}\*{\beta}^T\*{\beta} +C_p\*{1}_p^T{[\*{1}_p-(\*{X}_{[p]}\*{\beta}+\*{1}_p\beta_0)]}_+\\ 
&+C_u{[\*{1}_u+(\*{X}_{[u]}\*{\beta}+\*{1}_u\beta_0)]^T}[\*{1}_u+(\*{X}_{[u]}\*{\beta}+\*{1}_u\beta_0)]\\
&+{(\*{X}_{[ pu]} \*{\beta}+\*{1}_{ pu}\beta_0)^T\*R(\*{X}_{[ pu]} \*{\beta}+\*{1}_{ pu}\beta_0)}.
\end{aligned}
\end{equation}



However, the hinge loss term $\*{1}_p^T{[\*{1}_p-(\*{X}_{[p]}\*{\beta}+\*{1}_p\beta_0)]}_+$ in the objective function of PUAL in Equation (\ref{opt}) is not always differentiable in the feasible region of the optimization, bringing difficulty to  applying the gradient descent directly. To find an alternative way for solving PUAL, the following reformulation of Equation (\ref{opt}) can be considered:
\begin{equation}\label{admm}
\begin{aligned}
\min\limits_{\*{\beta},\beta_0,\*{h}}&
C_p\*{1}_p^T{[\*{h}]}_++C_u(\*{1}_u+\*{X}_{[u]}\*{\beta}+\*{1}_u\beta_0)^T(\*{1}_u+\*{X}_{[u]}\*{\beta}+\*{1}_u\beta_0)\\
+&(\*{X}_{[ pu]} \*{\beta}+\*{1}_{ pu}\beta_0)^T\*{R}(\*{X}_{[ pu]} \*{\beta}+\*{1}_{ pu}\beta_0) +\frac{\lambda}{2}\*{\beta}^T\*{\beta} \\
&s.t.\ \*{h}=\*{1}_p-(\*{X}_{[p]}\*{\beta}+\*{1}_p\beta_0).
\end{aligned}
\end{equation}

The predictive score function of PUAL for instance $\*x$ is the same as the predictive score function of SVM in Equation (\ref{prebsvm}).

 \subsubsection{Parameter Estimation}
 The convex objective function in Equation (\ref{admm})  can be regarded as the sum of the functions of $(\*{\beta},\beta_0)$ and the function of $\*{h}$ while the constraints in Equation (\ref{admm}) can be regarded as a linear combination of $(\*{\beta},\beta_0)$ and  $\*{h}$; this meets the requirement of ADMM~\cite{admm0,boyd2011distributed}, which can decompose a large-scale convex optimization problem with affine constraints into several simpler sub-problems and update the solution iteratively until convergence.  Moreover, ADMM is able to converge to modest accuracy within fewer iterations than the gradient descent. Hence, we adopt it here.

The Lagrangian function of problem in Equation (\ref{admm}) is
\begin{equation}\label{lar}
\begin{aligned}
\mathcal{L}(\*{\theta})=&C_p\*{1}_p^T{[\*{h}]}_++C_u(\*{1}_u+\*{X}_{[u]}\*{\beta}+\*{1}_u\beta_0)^T(\*{1}_u+\*{X}_{[u]}\*{\beta}+\*{1}_u\beta_0)\\+&(\*{X}_{[ pu]} \*{\beta}+\*{1}_{ pu}\beta_0)^T\*R(\*{X}_{[ pu]} \*{\beta}+\*{1}_{ pu}\beta_0) +\frac{\lambda}{2}\*{\beta}^T\*{\beta}\\
&+\*{u_h}^T[\*{1}_p-(\*{X}_{[p]}\*{\beta}+\*{1}_p\beta_0)-\*{h}],
\end{aligned}
\end{equation}
where $\*\theta=\{\*{\beta},\beta_0,\*{h},\*{u_h}\}$ and $\*{u_h}$ is dual variable. Then the augmented Lagrangian function is given as
\begin{equation}\label{key}
\begin{aligned}
\mathcal{L}_a(\*\theta)=&\mathcal{L}(\*{\theta})+\frac{{\mu_1}}{2}\left \|\*{1}_p-(\*{X}_{[p]}\*{\beta}+\*{1}_p\beta_0)-\*{h}\right \|_2^2,
\end{aligned}
\end{equation}
where $\mu_1$ is the step-size coefficient of the augmented Lagrangian function.

It follows that the update of $\*{\beta}$ and $\beta_0$ is 
\begin{equation}\label{beta}
\begin{aligned}
(\*{\beta}^{(k+1)},\beta_0^{(k+1)})&=\arg\min\limits_{\*{\beta},\beta_0}\frac{\lambda}{2} \*{\beta}^T\*{\beta}\\
&+C_u(\*{1}_u+\*{X}_{[u]}\*{\beta}+\*{1}_u\beta_0)^T(\*{1}_u+\*{X}_{[u]}\*{\beta}+\*{1}_u\beta_0)\\
&+(\*{X}_{[ pu]} \*{\beta}+\*{1}_{ pu}\beta_0)^T\*R(\*{X}_{[ pu]} \*{\beta}+\*{1}_{ pu}\beta_0)\\
&+{{\*{u_h}}^{(k)}}^T[\*{1}_p-(\*{X}_{[p]}\*{\beta}+\*{1}_p\beta_0)-\*{h}^{(k)}]\\
&+\frac{{\mu_1}}{2}\left \|\*{1}_p-(\*{X}_{[p]}\*{\beta}+\*{1}_p\beta_0)-\*{h}^{(k)}\right \|_2^2,
\end{aligned}
\end{equation}
which is a quadratic optimization with every term differentiable. 
Let $\*{I}_{k}, \forall k \in \mathbb{Z}$,  denote a $k\times k$ identity matrix. By defining 
\begin{equation}
\label{mat_beta}
\begin{aligned}
&\*{M}_{11}=\lambda\*{I}_{m} +2C_u\*{X}_{[u]}^T\*{X}_{[u]} +2\*{X}_{[ pu]}^T\*R\*{X}_{[ pu]}+\mu_1\*{X}_{[p]}^T\*{X}_{[p]},
\\&\*{M}_{12}=2C_u\*{X}_{[u]}^T\*{1}_{u}+2\*{X}_{[ pu]}^T\*R\*{1}_{ pu}+\mu_1\*{X}_{[p]}^T\*{1}_p,\\
&\*{M}_{21}=2C_u\*{1}_u^T\*{X}_{[u]}+2\*{1}_{ pu}^T\*R\*{X}_{[ pu]}+\mu_1\*{1}^T_p\*{X}_{[p]},\\&
M_{22}=2C_u n_u+2\*{1}^T_{ pu}\*R\*{1}_{ pu}+\mu_1n_p,\\
&\*{m}_{1}=-2C_u\*{X}^T_{[u]}\*{1}_u+\*{X}_{[p]}^T\*{u}_{\*{h}}+\mu_1\*{X}_{[p]}^T(\*{1}_{p}-\*{h}),\\
&m_{2}=-2C_un_u+\*{u}_{\*{h}}^T\*{1}_{p}+\mu_1{( \*{1}_{p}-\*{h} )}^T\*{1}_p,
\end{aligned}
\end{equation}
the solution of problem in Equation (\ref{beta}) can be obtained by solving the following linear equation w.r.t. $\*{\beta}$ and $\beta_0$:
\begin{equation}\label{betasolution}
\begin{aligned}
\begin{bmatrix}
\*{M}_{11}&\*{M}_{12} \\ 
\*{M}_{21}&M_{22} 
\end{bmatrix}
\begin{bmatrix}
\*{\beta}^{(k+1)}\\\beta_0^{(k+1)}
\end{bmatrix}
=
\begin{bmatrix}
\*{m}_{1}\\m_{2}
\end{bmatrix}.
\end{aligned}
\end{equation}

Then the update of $\*{h}$ becomes 
\begin{equation}\label{h}
\begin{aligned}
\*{h}^{(k+1)}=&\arg\min\limits_{\*{h}}C_p\*{1}_p^T{[\*{h}]}_++{\*{u_h}^{(k)}}^T[\*{1}_p-(\*{X}_{[p]}\*{\beta}^{(k+1)}+\*{1}_p\beta_0^{(k+1)})-\*{h}]\\
+&\frac{\mu_1}{2}\left \|\*{1}_p-(\*{X}_{[p]}\*{\beta}^{(k+1)} +\*{1}_p\beta_0^{(k+1)})-\*{h}\right \|_2^2,
\end{aligned}
\end{equation}
which is equivalent to solving the problem
\begin{equation}\label{h2}
\min\limits_{\*{h}}\sum_{i=1}^{n_p}{\left\{\frac{C_p}{\mu_1}[h_i]_++\frac{1}{2}{[1+\frac{u_{\*{h}i}^{(k)}}{\mu_1}-(\*{x}_{i}^T\*{\beta}^{(k+1)} +\beta_0^{(k+1)})-h_i]^2}\right\}}.
\end{equation}
According to~\cite{ye2011split}, for constant $c>0$, we can obtain 
\begin{equation}
\arg\min\limits_{x}c[x]_++\frac{1}{2}\|x-d\|^2_2=\begin{cases}
d-c & d>c,\\
0 & 0 \leq d \leq c,\\
d & d<0.
\end{cases}
\end{equation}
Thus, by defining $s_c(d)=\arg\min\limits_{x}c[x]_++\frac{1}{2}\|x-d\|^2_2$, the $i$th element of $\*{h}^{(k+1)}$ in problem in Equation (\ref{h}) is solved as
\begin{equation}\label{h3}
h_i^{(k+1)}=s_{\frac{C_p}{\mu_1}}{\left(1+\frac{u_{\*{h}i}^{(k)}}{\mu_1}-(\*{x}_{i}^T\*{\beta}^{(k+1)} +\beta_0^{(k+1)})\right)}, i=1, \dots, n_p.
\end{equation}

According to \cite{boyd2011distributed}, the update of the dual parameter $\*u_h$ can be 
\begin{equation}
     \*{u_h}^{(k+1)}=\*{u_h}^{(k)}+\mu_1[\*{1}_p-(\*{X}_{[p]}\*{\beta}^{(k+1)}+\*{1}_p\beta_0^{(k+1)})-\*{h}^{(k+1)}].
\end{equation}

\subsubsection{Algorithm}
The algorithm of PUAL can be summarized in Algorithm~\ref{al1}.

 \begin{algorithm}
        \caption{PUAL with linear decision boundary}
         \label{al1}
        \hspace*{0.02in} {\bf Input:} PU dataset, $C_p$, $C_u$, $\lambda$, $\sigma$ and $\mu_1$\\ 
        \hspace*{0.02in} {\bf Output:} $\*\beta$ and $\beta_0$
        \begin{algorithmic}[1] 
        \State Initialize $\*\beta$, $\beta_0$, $\*h$, $\*u_h$
        
        \While {not converged}
        \State Update $(\*{\beta}^{(k+1)},\beta_0^{(k+1)})=\arg\min\limits_{\*{\beta},\beta_0}\mathcal{L}_a(\*{\beta},\beta_0,\*{h}^{(k)},\*{u_h}^{(k)})$
        \State Update  $\*{h}^{(k+1)}=\arg\min\limits_{\*{h}}\mathcal{L}_a(\*{\beta}^{(k+1)},\beta_0^{(k+1)},\*{h},\*{u_h}^{(k)})$
        \State Update $\*{u_h}^{(k+1)}=\*{u_h}^{(k)}+\mu_1[\*{1}_p-(\*{X}_{[p]}\*{\beta}^{(k+1)}+\*{1}_p\beta_0^{(k+1)})-\*{h}^{(k+1)}]$
        \EndWhile

        \end{algorithmic}
    \end{algorithm}

\subsection{PUAL with Non-Linear Decision Boundary}
\label{sec:kernel}

In this section, we develop a kernel-based algorithm to enable PUAL to have non-linear decision boundary, so that PUAL can be applied on the non-linear separable datasets including trifurcate PU datasets. The techniques applied in this section are similar to many previous methods~\cite{ke2018global,jain2016nonparametric,christoffel2016class,bekker2018estimating}.

\subsubsection{Objective Function}
  Suppose $\*\phi(\*x) \in \mathbb{R}^{ r \times 1}$ be a mapping of the instance vector $\*x$. Then let $\*\phi(\*X_{[k]}) \in \mathbb{R}^{n_k\times r},k=p,u, pu$ be the mapping of the original data matrix $\*X_{[k]}$. The $i$th row of $\*\phi(\*X_{[k]})$ is $\phi(\*x_i)^T$.    According to Equations (\ref{mat_beta}) and (\ref{betasolution}), using $\*\phi(\*X_{[k]})$ as features matrix instead of $\*X_{[k]}$ for the training of PUAL, we can find the following necessary condition for the optimal solution of $\*\beta$: 
\begin{equation}
\label{k_rel0}
\begin{aligned}
&\*B\*{\beta}^{}=\*\phi(\*X_{[ pu]})^T\*\Omega^{},
\end{aligned}
\end{equation}
where
\begin{equation}
 \label{matb}
\*B=\*{M}_{11}-\dfrac{\*{M}_{12}\*{M}_{21}}{M_{22}},
\end{equation}
and
\begin{equation}
\label{upomega}
\*{\Omega}^{}=\begin{bmatrix}
\*{u}_{\*{h}}^{}-\mu_1\frac{m_2}{M_{22}}\*{1}_{p}+\mu_1(\*{1}_{p}-\*{h}^{})\\
-2C_u\*{1}_{u}-2\frac{m_2}{M_{22}}C_u\*{1}_{u}
\end{bmatrix}-2\frac{m_2}{M_{22}}\*R\*{1}_{ pu}.
\end{equation}

Equation (\ref{k_rel0}) can be regarded as a condition when the objective function reaches its minimum. For $n_p,n_u>m$, $\*B$ is symmetric and invertible. In this case, we can obtain
\begin{equation}
\label{k_rel}
\*{\beta}=\*B^{-1}\*\phi(\*X_{[ pu]})^T\*\Omega.
\end{equation}
Substituting Equation (\ref{k_rel}) into the objective function in Equation (\ref{opt}), we have the new objective function as
\begin{equation}
\begin{aligned}
 \label{ffar}
\min\limits_{\*\Omega,\beta_0}&\frac{\lambda}{2} \*\Omega^T\*\phi(\*X_{[ pu]})\*B^{-1}\*B^{-1}\*\phi(\*X_{[ pu]})^T\*\Omega +C_p\*{1}_p^T{[\*{1}_p-(\*\phi(\*X_{[p]})\*B^{-1}\*\phi(\*X_{[ pu]})^T\*\Omega+\*{1}_p\beta_0)]}_+\\
+&C_u{[\*{1}_u+\*\phi(\*X_{[u]})\*B^{-1}\*\phi(\*X_{[ pu]})^T\*\Omega+\*{1}_u\beta_0]}{[\*{1}_u+\*\phi(\*X_{[u]})\*B^{-1}\*\phi(\*X_{[ pu]})^T\*\Omega+\*{1}_{u}\beta_0]^T}\\
+&(\*\phi(\*X_{[ pu]})\*B^{-1}\*\phi(\*X_{[ pu]})^T\*\Omega+\*{1}_{ pu}\beta_0)^T
\*R(\*\phi(\*X_{[ pu]})\*B^{-1}\*\phi(\*X_{[ pu]})^T\*\Omega+\*{1}_{ pu}\beta_0).
\end{aligned}
\end{equation}

To prove $\*\phi(\*X_{[ k]})\*B^{-1}\*\phi(\*X_{[ pu]})^T$ and $\*\phi(\*X_{[ k]})\*B^{-1}\*B^{-1}\*\phi(\*X_{[ pu]})^T$ in Equation (\ref{ffar}) are kernel matrices for $\*X_{[ k]}$ and $\*X_{[ pu]}$,   we need to introduce the two closure properties for the construction of kernel functions proved in~\cite{genton2001classes} as follows.
     \begin{theorem}
  \label{closure}
Let $\phi(\*X),\phi(\*Z)$ be a mapping of  matrices of $\*X, \*Z$  and $\*\kappa_1(\phi(\*X),\phi(\*Z))$ be a  kernel matrix of $\phi(\*X)$ and $\phi(\*Z)$. Then the following two matrices  $\*\kappa_2(\mathbf{\*X}, \mathbf{\*Z})$ and  $\*\kappa_3(\mathbf{\*X}, \mathbf{\*Z})$ can be regarded as the kernel matrix w.r.t. $\*X, \*Z$:
        \begin{itemize}
        \item $\*\kappa_2(\mathbf{\*X}, \mathbf{\*Z})=\*\kappa_1(\phi(\mathbf{\*X}), \phi(\mathbf{\*Z})),$
\item $\*\kappa_3(\mathbf{\*X}, \mathbf{\*Z})=\mathbf{\*X} \*F \mathbf{\*Z}^T,$ with $\*F$  to be a symmetric matrix.
        \end{itemize}
   \end{theorem}
Then, according to the closure properties in Theorem \ref{closure}, we can obtain
\begin{equation}
 \label{ck1}
\begin{aligned}
\*\phi(\*X_{[k]})\*B^{-1}\*\phi(\*X_{[ pu]})^T=&\*\Phi'(\*\phi(\*X_{[k]}),\*\phi(\*X_{[ pu]}))\\
=&\*\Phi(\*X_{[k]},\*X_{[ pu]})
\end{aligned}
\end{equation}
and
\begin{equation}
  \label{ck2}
\begin{aligned}
\*\phi(\*X_{[k]})\*B^{-1}\*B^{-1}\*\phi(\*X_{[ pu]})^T=&\*\Phi''(\*\phi(\*X_{[k]}),\*\phi(\*X_{[ pu]}))\\
=&\*\Phi_2(\*X_{[k]},\*X_{[ pu]}),
\end{aligned} 
\end{equation}
where $\*\Phi'(\*\phi(\*X_{[k]}),\*\phi(\*X_{[ pu]}))$, $\*\Phi''(\*\phi(\*X_{[k]}),\*\phi(\*X_{[ pu]}))$ are the kernel matrices for $\*\phi(\*X_{[k]})$ and $\*\phi(\*X_{[pu]})$, and $\*\Phi(\*X_{[k]},\*X_{[ pu]}), \*\Phi_2(\*X_{[k]},\*X_{[pu]})$ are the kernel matrices for $\*X_{[k]}$ and $\*X_{[pu]}$.

Hence, the objective function of PUAL can be reformulated as
\begin{equation}
 \label{kob1}
\begin{aligned}
\min\limits_{\*\Omega,\beta_0}&\frac{\lambda}{2} \*\Omega^T\*\Phi_2(\*X_{[ pu]},\*X_{[ pu]})\*\Omega +C_p\*{1}_p^T{[\*{1}_p-(\*\Phi(\*X_{[p]},\*X_{[ pu]})\*\Omega+\*{1}_p\beta_0)]}_+\\
+&C_u{[\*{1}_u+\*\Phi(\*X_{[u]},\*X_{[ pu]})\*\Omega+\*{1}_u\beta_0]^T}{[\*{1}_u+\*\Phi(\*X_{[u]},\*X_{[ pu]})\*\Omega+\*{1}_{u}\beta_0]}\\
+&{(\*\Phi(\*X_{[ pu]},\*X_{[ pu]})\*\Omega+\*{1}_{ pu}\beta_0)^T\*R(\*\Phi(\*X_{[ pu]},\*X_{[ pu]})\*\Omega+\*{1}_{ pu}\beta_0)},
\end{aligned}
\end{equation}
whose solution is not related to $\*X_[k],k=p.u.pu$ once the kernel  matrices are determined.

The predictive score function for instance $\*x^*$ of PUAL can be now transformed to 
\begin{equation}
f=\*\Phi(\*x^*,\*X_{[ pu]})\*\Omega+\beta_0.
\end{equation}

 \subsubsection{Parameter Estimation}
 
In this case, we can update $\beta_0$ via
\begin{equation}
\label{betaaa_0}
\beta_0^{(k+1)}=\frac{m_2}{M_{22}}-\*Q_b^{(k+1)}/M_{22},
\end{equation}
where $m_2$, $M_{22}$ are not related to $\*X_{[p]},\*X_{[u]},\*X_{[ pu]}$ and
\begin{equation}
\begin{aligned}
\*Q_b^{(k+1)}=&2C_u\*{1}_{u}^T\*\Phi(\*X_{[u]},\*X_{[ pu]})\*\Omega^{(k+1)}\\
+ &2\*{1}_{ pu}^T\*R\*\Phi(\*X_{[ pu]},\*X_{[ pu]})\*\Omega^{(k+1)}\\
+ &\mu_1\*{1}_{p}^T\*\Phi(\*X_{[p]},\*X_{[ pu]})\*\Omega^{(k+1)}.
\end{aligned}
\end{equation}

The update of $\*h, \*u_{\*{h}}$ can be reformulated as 
\begin{equation}
\label{upnew}
\begin{aligned}
&\*h_i^{(k+1)}=s_{\frac{C_p}{\mu_1}}{\left(1+\frac{u_{\*{h}i}^{(k)}}{\mu_1}-(\*\Phi(\*x_i,\*X_{[ pu]})\*\Omega^{(k+1)} +\beta_0^{(k+1)})\right)}, i=1,\dots,n_p,
\\&\*{u_h}^{(k+1)}=\*{u_h}^{(k)}+\mu_1[\*{1}_p-(\*\Phi(\*X_{[p]},\*X_{[ pu]})\*\Omega^{(k+1)}+\*{1}_p\beta_0^{(k+1)})-\*{h}^{(k+1)}].
\end{aligned}
\end{equation}

As $\*\Phi_2(\*X_{[k]},\*X_{[ pu]})$ does not directly appear in the update process for the optimization, we only need to determine the form of $\*\Phi(\*X_{[k]},\*X_{[ pu]})$ in practice. Moreover,  $\lambda$  either does not appear directly in the above  update process or it is contained in the matrix $\*B$ as a part of $\*\Phi(\*X_{[k]},\*X_{[ pu]})$. Therefore, for convenience, in the case of using the kernel trick in Section \ref{experiments321}, we use $\lambda$  to represent the hyper-parameter(s) of the kernel matrix $\*\Phi(\*X_{[k]},\*X_{[ pu]})$.

\subsubsection{Algorithm}
The algorithm of PUAL with non-linear  decision boundary can be summarized in Algorithm~\ref{al2}.

 \begin{algorithm}
        \caption{PUAL with non-linear decision boundary}
         \label{al2}
        \hspace*{0.02in} {\bf Input:} PU dataset, $\*\Phi$, $C_p$, $C_u$, $\lambda$, $\sigma$  and $\mu_1$\\
        \hspace*{0.02in} {\bf Output:} $\*\Omega$ and $\beta_0$
        \begin{algorithmic}[1]
        \State Initialize $\*\Omega$, $\beta_0$, $\*h$, $\*u_h$.
        \While {not converged}
        \State Update $\*\Omega$ via Equation (\ref{upomega}) w.r.t. $\*h^{(k)}$ and $\*{u_h}^{(k)}$
        \State Update $\beta_0$ via Equation (\ref{betaaa_0})
        \State Update $\*h$, $\*u_{\*h}$ via Equation (\ref{upnew})
        \EndWhile

        \end{algorithmic}
    \end{algorithm}

\section{Experiments}
\label{EXP}
\subsection{Experiments on Synthetic Data}
\label{experiments321}
In this section, we conduct experiments on  synthetic datasets following the pattern in  Fig.~\ref{demo_gllc} to verify the superior performance of PUAL over GLLC.
\subsubsection{Generation of Synthetic Positive-Negative (PN) Datasets}
\label{gens}
The synthetic datasets following the pattern in Fig.~\ref{demo_gllc} were obtained by the following steps:
\begin{enumerate}
\item To generate the first subset of the  2-dimensional synthetic positive set, 200 instances were sampled from the multivariate normal distribution with  mean vector $(15,15)$ and the covariance matrix
$    \begin{bmatrix}
50 & 0\\ 
 0& 50
\end{bmatrix}.
$
\item  To generate the second subset of the 2-dimensional synthetic positive set, 200 instances were sampled from the multivariate normal distribution with the  mean vector $(\mathbf{mean}_{p2},\mathbf{mean}_{p2})$ and the covariance matrix
$    \begin{bmatrix}
$50$ & 0\\ 
 0& $50$
\end{bmatrix}.
$

\item To generate the 2-dimensional synthetic negative set, 400 instances were sampled from the multivariate normal distribution of mean vector $(0,0)$ and the covariance matrix
$    \begin{bmatrix}
50 & 0.2\\ 
 0.2& 50
\end{bmatrix}.
$

\item Mixing the first subset of the synthetic positive set obtained in Step 1, the second subset of the synthetic positive set obtained in Step 2  and the synthetic negative set obtained in Step 3, a simple 2-dimensional synthetic dataset can be eventually obtained as shown in Fig.~\ref{demo_gllc}.
\end{enumerate}

 In Step 2, $\mathbf{mean}_{p2}$ took value from $(50,100,200,500,1000)$. For each value of $\mathbf{mean}_{p2}$, the above steps were repeated 5 times so that we have 5 synthetic datasets for each value of $\mathbf{mean}_{p2}$.

\subsubsection{Training-Test Split for the Synthetic PU   Datasets}
\label{split}
It should be  noted that both GLLC and PUAL can be applied on the datasets sampled from either single‑training‑set scenario~\cite{bekker2020learning}  or case-control  scenario~\cite{du2014analysis} as we set the suitable metric for hyper-parameter tuning in practice. More specifically,  the case-control scenario indicates that the unlabeled training set can be regarded to be i.i.d. sampled from the population, while the single‑training‑set  scenario indicates that the whole training set can be regarded to be i.i.d. sampled from the population. In this case, for more intuitive comparison, we split each of the synthetic dataset generated in Section \ref{gens} to construct the PU training and test sets consistent with the single‑training‑set scenario by the following two steps:
\begin{enumerate}
\item Firstly, to split the dataset into a training set and a test set, 70\% of the instances in the simple synthetic dataset obtained in Section \ref{gens} were randomly selected as the training set while the test set was constituted by the rest 30\%  instances.

\item Secondly, to construct the labeled-positive set and unlabeled-set for training, 25\% of the positive instances in the above obtained training set were randomly selected to form the labeled-positive set for training. The rest of the positive instances were mixed with the negative set, contributing to the unlabeled set for training.
\end{enumerate}
Then 25 pairs of PU training set and test set were obtained. 

\subsubsection{Model Setting}
\label{orms1}
We note that the real value of F1-score on the training dataset is not accessible if we do not use the label information during model training. Therefore, by fixing $C_p$ to 1 and the number $K$ of the nearest neighbors to $5$, $C_u$, $\lambda$, $\sigma$  in the objective functions of PUAL and GLLC were determined by 4-fold cross-validation (CV), which reached the highest average PUF-score  proposed in~\cite{lee2003learning} on the validation sets. PUF-score is similar to F1-score and can be directly obtained from PU data:
\begin{equation}
 \label{PUF}
\mathrm{\text{PUF-score}} =\frac{\text{recall}^2}{P[\text{sgn}(f(\*x))=1]},
\end{equation}
where `$\text{recall}$' can be estimated by computing $\frac{1}{n_p} \sum_{\*x_i \in X_{[p]}} \mathbb{i}(\text{sgn}(f(\*x_i))=1)$ with the indicator function denoted by $\mathbb{i}(\cdot)$, and at the single‑training‑set scenario $P[\text{sgn}(f(\*x))=1]$ can be estimated by computing $\frac{1}{n_{p}+n_{u}}\sum_{\*x_i \in  X_{[pu]}}\mathbb{i}(\text{sgn}(f(\*x_i))=1)$. Furthermore, $\lambda$, $\sigma$ were tuned from the set $\{1,2,3,4,5\}\circ \{0.1,1,10,100\}$; $C_u$ was selected from the set $\{0.01,0.02,\dots, 0.5\}$ based on the setting in~\cite{ke2018global}.

\subsubsection{Results and Analysis}
\begin{figure}[htbp]
\centering
{\includegraphics[width=5.5cm, angle=0]{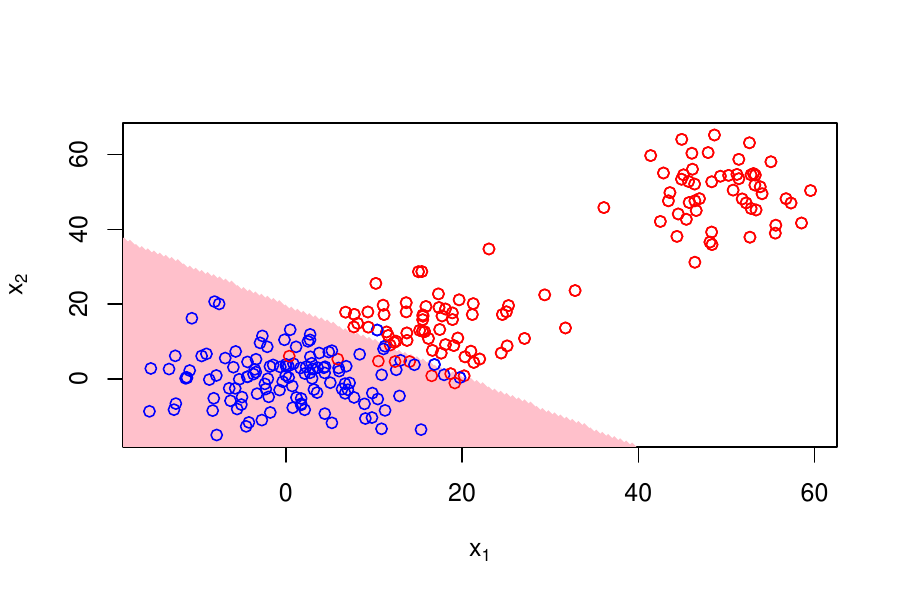}}
{\includegraphics[width=5.5cm, angle=0]{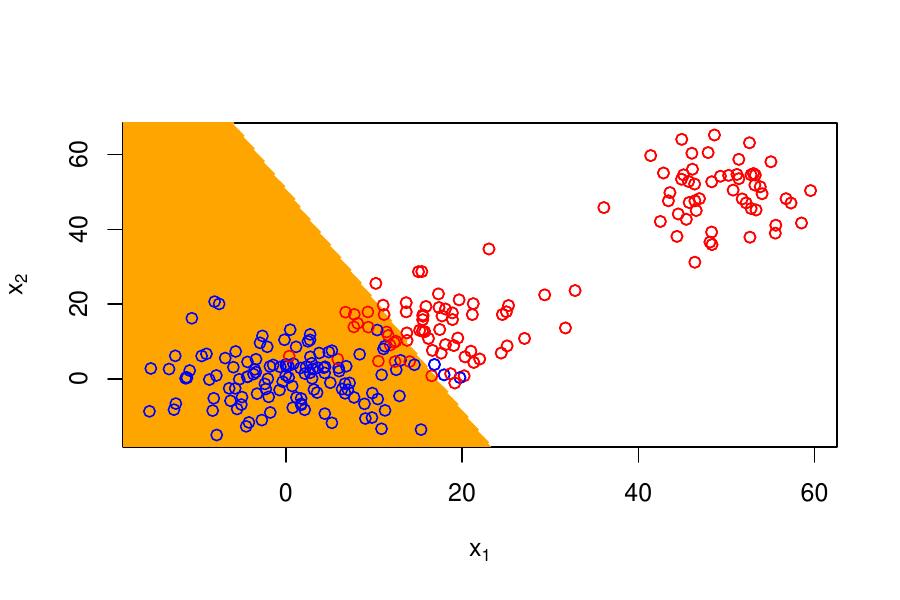}}
 {\includegraphics[width=5.5cm, angle=0]{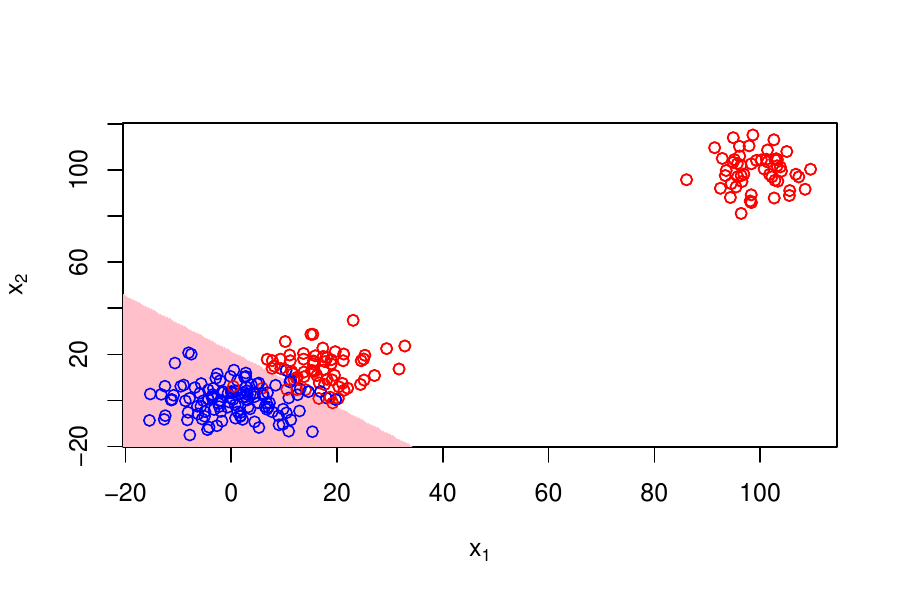}}
{\includegraphics[width=5.5cm, angle=0]{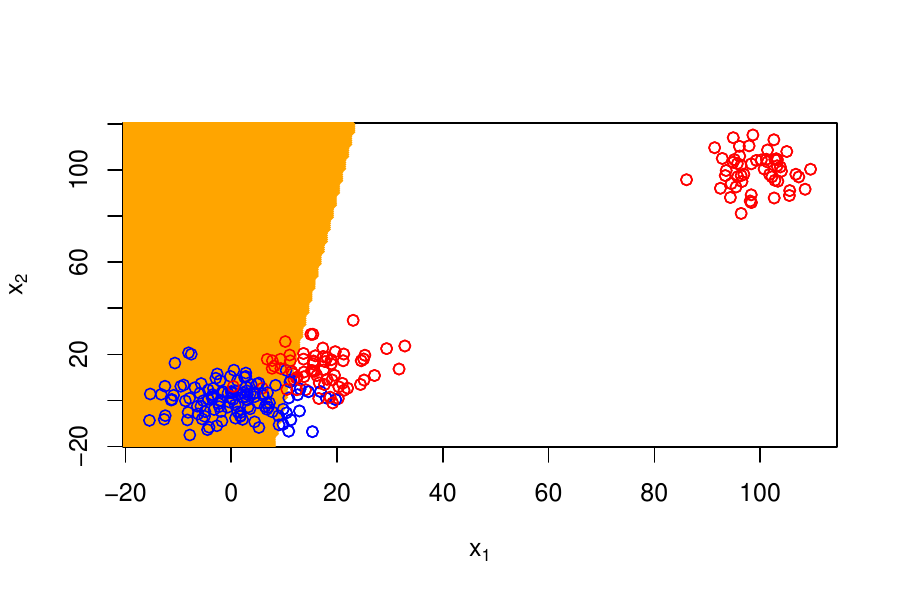}}

{\includegraphics[width=5.5cm, angle=0]{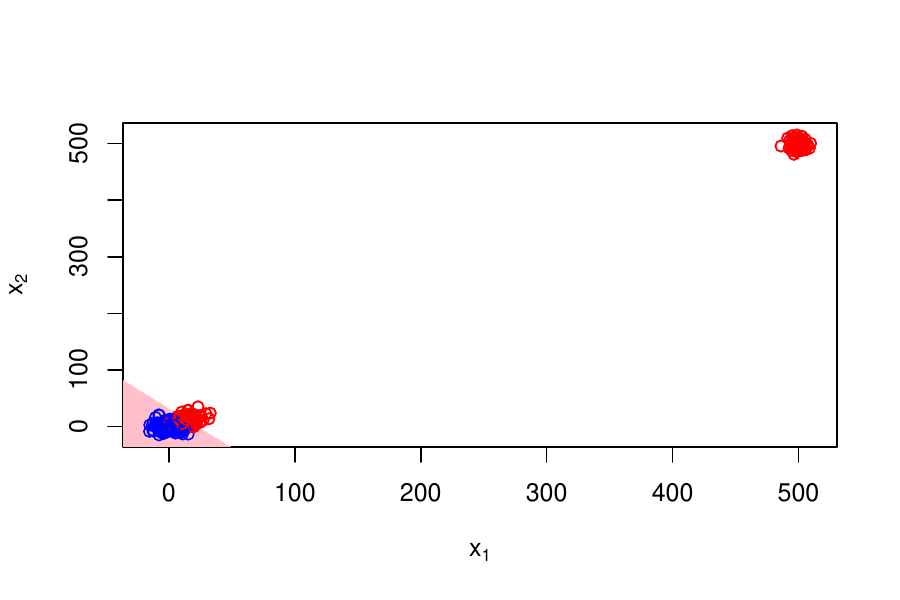}}
{\includegraphics[width=5.5cm, angle=0]{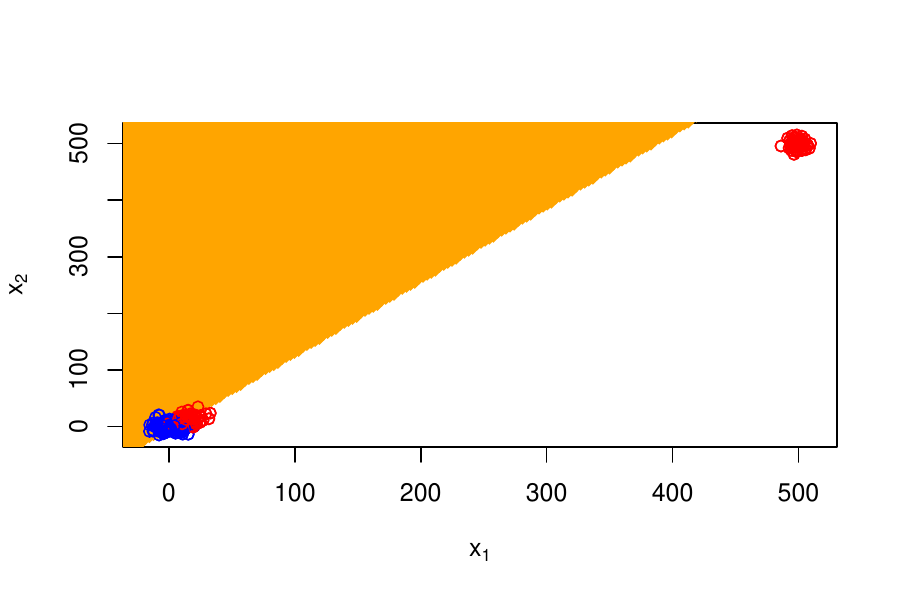}}

\caption{The decision boundaries trained by (left) PUAL and (right) GLLC on the synthetic data with ${\mathbf{mean}_{p2}}=50, 100, 500$. Pink area: the negative region of PUAL; orange area: the negative region of GLLC; red points: positive instances; blue points: negative instances; the instances in the plots are from the test sets.}
  \label{synth1}
\end{figure}

Results of the experiments on the constructed synthetic PU datasets  are  summarized in Table \ref{synre}. The performance are measured by the average F1-score, which is a widely-used metric for the evaluation of PU learning methods~\cite{liu2023robust}.  The patterns of the decision boundary obtained by PUAL and GLLC on the synthetic datasets are illustrated in Fig.~\ref{synth1}. For each value of ${\mathbf{mean}_{p2}}$, we use the first generated synthetic dataset as example.

\begin{table}[htbp]
  \caption{Summary of the average  F1-score (\%) with the standard deviation, from the experiments on the synthetic dataset; the result highlighted in blue is the better one between PUAL and GLLC.}
   \begin{tabular}{r|rr}
    \toprule
    $\mathbf{mean}_{p2}$     & \multicolumn{1}{l}{PUAL} & \multicolumn{1}{l}{GLLC} \\
    \midrule
    50    & \textcolor{blue}{95.07 $\pm$ 0.78}& 91.17 $\pm$ 1.52\\
    100   & \textcolor{blue}{94.89 $\pm$ 0.61}& 86.27 $\pm$ 1.83\\
    200   & \textcolor{blue}{93.56 $\pm$ 0.75}& 81.04 $\pm$ 2.93\\
    500   & \textcolor{blue}{92.83 $\pm$ 3.22}& 73.58 $\pm$ 2.58\\
    1000  & \textcolor{blue}{93.47 $\pm$ 2.08}& 71.28 $\pm$ 2.37\\
    \bottomrule
    \end{tabular}%
  \label{synre}
\end{table}%

According to the experimental results in Table \ref{synre}, PUAL always has better performance than GLLC on the synthetic PU datasets with all the 5 values of ${\mathbf{mean}_{p2}}$. Furthermore, with the value of ${\mathbf{mean}_{p2}}$  increasing, the gap between PUAL and GLLC becomes increasingly large. This is more clearly in the six plots in Fig.~\ref{synth1}, as one of the positive subset becomes increasingly far away, the decision boundary of GLLC is dragged to a strange position that more and more instances are misclassified by the decision boundary of GLLC, while the decision boundary of PUAL is unaffected. Therefore, it is verified that PUAL can generate much better linear decision boundary   than GLLC on the datasets following the pattern in Fig.~\ref{demo_gllc}.

\subsection{Experiments on Real-World Data}\label{experiments123}

In this section, we further assess the classification performance of PUAL on real-world datasets.

\subsubsection{Real-World  Datasets}
\label{data1321}
Firstly, 16 datasets from the UCI Machine Learning Repository (https://archive.ics.uci.edu/ml/index.php) were selected to assess the performance of PUAL and verify our motivation: Accelerometer (\textbf{Acc}), \textbf{Ecoli}, Pen-Based Recognition of Handwritten Digits (\textbf{Pen}), Online Retail (\textbf{OR1}), Online Retail II (\textbf{OR2}), Parking Birmingham (\textbf{PB}),  \textbf{wifi}, Sepsis survival minimal clinical records (\textbf{SSMCR}), Avila, Raisin Dataset (\textbf{RD}), Occupancy Detection (\textbf{OD}), User Knowledge Modeling Data Set (\textbf{UMD}), \textbf{Seeds}, Energy efficiency Data Set (\textbf{ENB}), Heart Disease (\textbf{HD}) and Liver Disorders (\textbf{LD}).  The details of these real-world datasets are summarized in Table \ref{tab:1}.

\begin{table}[htbp]
\caption{Summary of the real-world datasets.}
\label{tab:1}
\begin{tabular}{llll}
\hline
Dataset                                  & Positive instances & Negative instances & \# Features \\ \hline
\textbf{Acc}                            & 100 red          & 100 blue      & 4          \\
\textbf{Ecoli}                                    & 116 im \& 52 pp           & 143 cp  \&  25 om        & 6          \\
\textbf{Pen}                                     & 200 {one} \& 200 eight        & 400 four   & 16         \\
\textbf{OR1}                            & 301 UK      & 301 Germany         & 4          \\
\textbf{OR2}                         &  500 UK       &   500 Germany        & 4          \\
\textbf{SSMCR} & 391 alive       & 109 dead        & 3          \\
\textbf{PB}                       &500  Bull Ring & 500  BHMBCCMKT01  & 3         \\
\textbf{OD}                      & 100 occupied    & 300 not occupied& 5          \\
\textbf{UMD}         &  83  Low         &  63   high        & 5          \\
\textbf{Seeds}                                    & 70   Kama        & 70  Rosa         & 7          \\
\textbf{ENB}               & 144  TypeII      & 144 Type III     & 7          \\
\textbf{wifi}             & 100 Location 2\& 100 Location 4  & 499 Location 1 \& 100 Location 3  & 7          \\
    \textbf{Avila}                                    & 300 E           & 900 A           & 10          \\
\textbf{RD}                           & 450 Kecimen      & 450 Besni        & 7          \\
\textbf{LD}                         & 144 class 1      & 200 class 2      & 6          \\
\textbf{HD}                            & 150 absence  & 119 presence &13          \\
        \hline
\end{tabular}
\end{table}

\subsubsection{Compared Methods}
GLLC, uPU and nnPU were also trained on the 16 real datasets as the compared methods with PUAL. GLLC serves as the baseline of PUAL. uPU and nnPU are two consistent PU learning methods.

\subsubsection{Training-Test Split for the Real PU Datasets}
\label{sp_real}
Different from the steps in Section \ref{split}, PU training and test sets for the 16 real datasets are constructed under the case-control scenario since we would like to see the performance of PUAL on the datasets with various labeling mechanism. Furthermore, under the case-control scenario, it is possible to do fair comparison between GLLC and the two convincing methods for PU classification, i.e., uPU~\cite{du2014analysis} and nnPU~\cite{kiryo2017positive}, since they were proposed under the case-control scenario.  The steps to perform training-test split for the 16 PU real datasets are summarized as follows:
\begin{enumerate}
\item To obtain the binary positive-negative (PN) datasets from the original multi-class dataset, certain classes in each of the original real datasets were treated as positive while  some other classes were treated as negative with the rest of classes abandoned. 

\item To construct the labeled-positive set and unlabeled-set, $\gamma'$ of the positive instances in each binary PN dataset obtained in Step 1 were randomly selected to form the labeled-positive set, and the rest of the positive instances were mixed into the unlabeled set, contributing to the unlabeled set of the PU dataset.

\item To generate the training set and the test set from the constructed PU datasets, the labeled-positive set and $70\%$ of the instances in the unlabeled set were selected as the  training set while   the test set was constituted by the rest $30\%$ of the instances in the unlabeled set; this corresponds to the setting of case-control scenario since the unlabeled training set and the test set can be regarded to be sampled from the same population.
\end{enumerate}

During preliminary experiments, we found that  the label frequency $\gamma$ needs to be greater than $20\%$ for the hyper-parameter tuning strategy introduced in Section~\ref{ms} to achieve adequate results. Therefore, the value of $\gamma'$ is set to  $\frac{7}{17}$, $\frac{7}{37}$ so that we have the label frequency $\gamma=\gamma'/(0.3\gamma'+0.7)=0.5, 0.25$, which is the fraction of positive instances that are labeled, in the corresponding constructed PU training sets, respectively. Then Step 2 and Step 3 were repeated for 10 times on each of the 16 binary PN datasets and obtained 10 pairs of PU training and test sets for each of the 16 binary PN datasets with a certain value of $\gamma'$. 

\subsubsection{Model Setting}
\label{ms}
At the case-control scenario, we also use  PUF-score in Equation (\ref{PUF}) for hyper-parameter selection. The numerator `$\text{recall}$' can still be estimated by computing $\frac{1}{n_p} \sum_{\*x_i \in X_{[p]}} \mathbb{i}(\text{sgn}(f(\*x_i))=1)$, while the denominator   $P[\text{sgn}(f(\*x))=1]$ needs to be estimated by computing $\frac{1}{n_{u}}\sum_{\*x_i \in  X_{[u]}}\mathbb{i}(\text{sgn}(f(\*x_i))=1)$ at the case‑control scenario.  Therefore, by fixing $C_p$ to $1$ and the number  $K$ of the nearest neighbors to $5$,  $C_u$, $\lambda$, $\sigma$  in the objective functions of PUAL and GLLC were firstly tuned by 4-fold CV, which reached the highest average PUF-score.

Furthermore, considering the higher complexity of the real datasets compared with the synthetic datasets in Section \ref{experiments321}, we modified our strategy for hyper-parameter selection, enabling efficient selection of hyper-parameters across a broader range.  More specifically, $\lambda$, $\sigma$ were tuned from the set $\{10^{-4}, 10^{-3}, 10^{-2}, 10^{-1}, 10^{0}, 10^{1}, 10^{2},10^{3}, 10^{4}\}$ and   $C_u$ was  selected to from the set $\{0.5, 0.3, 0.1, 0.05, 0.01\}$  based on the setting in~\cite{ke2018global}. Then $\lambda$, $\sigma$ and $C_u$ were continually tuned following the greedy algorithm based on the  average PUF-score on the validation sets as follows:
\begin{enumerate}
    \item Set $\lambda$, $\sigma$ and $C_u$ to the best combination from the grid search.
    \item Sequentially update one of hyper-parameters $\lambda$, $\sigma$ and $C_u$ by increasing/decreasing 10\% of its current value with the rest of the hyper-parameters unchanged. The optimal scenario on 4-fold CV is set to be the final update of this step.
    \item Repeat Step 2 until there is no better scenario  appearing.
\end{enumerate}

In addition, the hyper-parameters of uPU and nnPU were fixed as the recommended setting in the open source provided by~\cite{kiryo2017positive} at \url{https://github.com/kiryor/nnPUlearning}. Radial Basis Function (RBF) kernel was applied to both PUAL and GLLC. More specifically, we computed $\exp \left( -{{\| x_i - x_j \|^2}}/{{2\lambda^2}} \right)$ as the $(i,j)$ element of $\*\Phi'(\*\phi(\*X_{[pu]}),\*\phi(\*X_{[ pu]}))$.

\subsubsection{Summary of Comparison between PUAL and GLLC}

\begin{figure}[htbp]    
\centering  
\subfloat[$\gamma=0.5$]{\label{fig:box_0.5}
\centering
\includegraphics[width=11cm, angle=0]{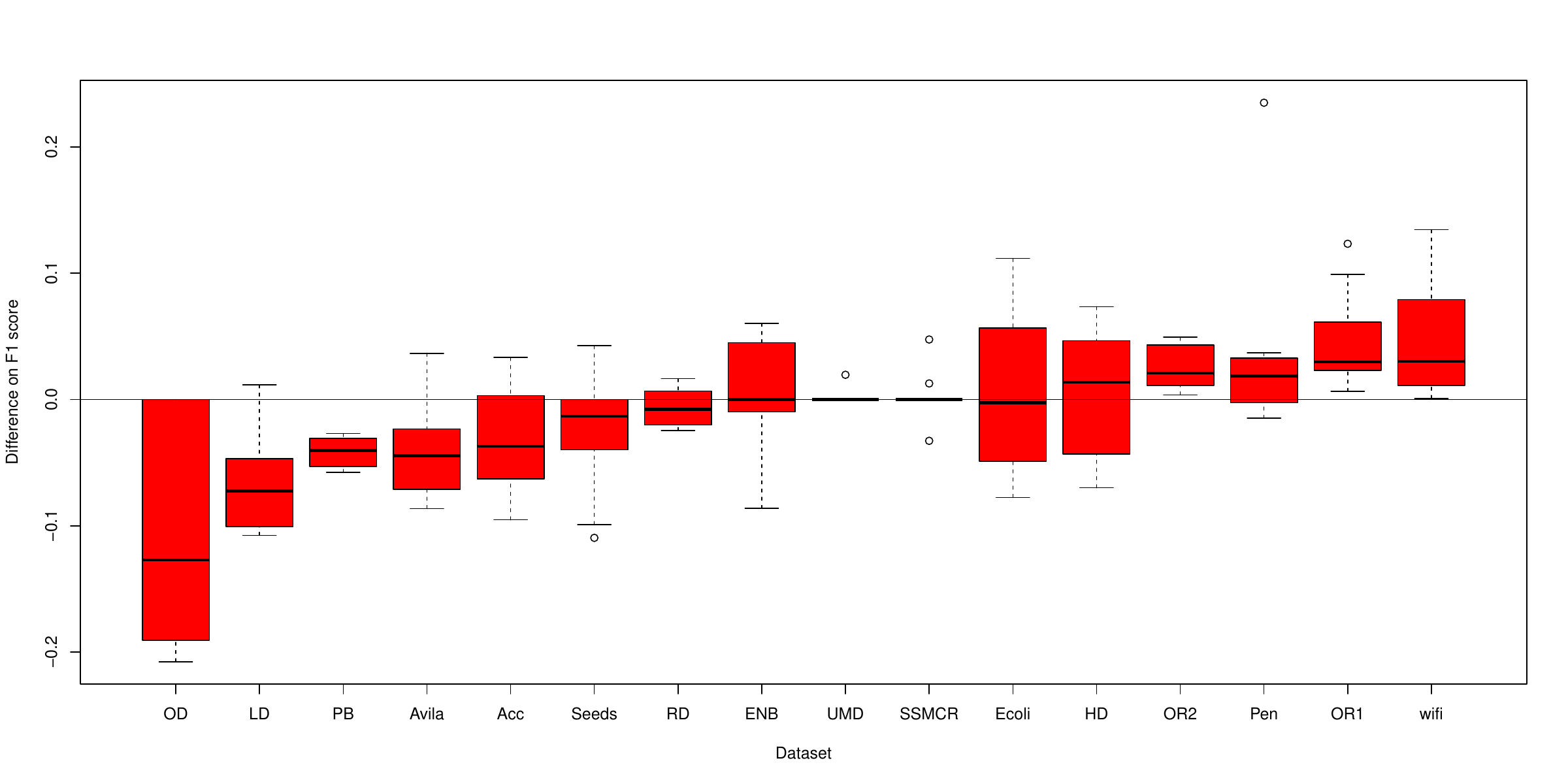}} 
\\
\subfloat[$\gamma=0.25$]{\label{fig:box_0.25}
\centering
\includegraphics[width=11cm, angle=0]{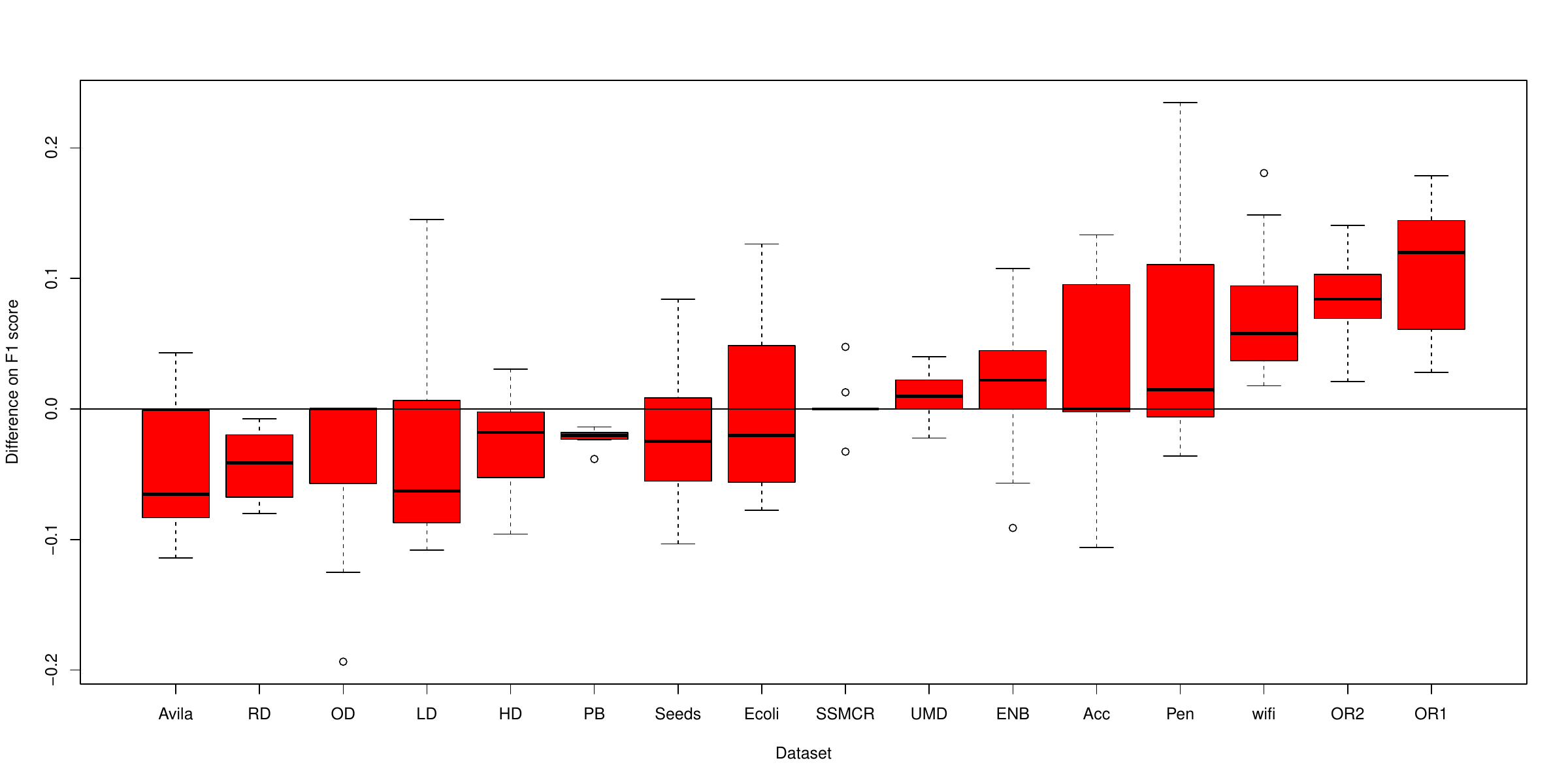}} 
\caption{Boxplots for the difference between F1-scores of PUAL and GLLC on each dataset increasingly ranked by medians; label frequencies $\gamma=0.5$ (top) and $0.25$ (bottom); x-axis: the datasets; y-axis: the difference between PUAL and GLLC in F1-score. }          
\label{box_0.25}
\end{figure}

The results of the difference between the F1-scores of PUAL and GLLC on each pairwise experiments are shown in the boxplots in Fig.~\ref{box_0.25} with $\gamma=0.5, 0.25$, respectively. In both figures,  the best four datasets for PUAL over GLLC are \textbf{wifi}, \textbf{OR1}, \textbf{OR2},  and \textbf{Pen}. Furthermore, with $\gamma=0.5, 0.25$, three of the  worst four datasets for PUAL compared with GLLC are \textbf{OD}, \textbf{LD}, \textbf{Avila}. When $\gamma=0.5$, the rest one of the worst four datasets is  \textbf{PB}; when $\gamma=0.25$, it is  \textbf{RD}. According to the t-tests with $p$-value lower than 0.05 between the pairwise results of these four methods, there are totally 8 cases where PUAL is the optimal choice among the four methods.

\subsubsection{Pattern Analysis for the Real Datasets Preferring PUAL to GLLC}
\label{rtog}

\begin{figure}[htbp]    
\centering           
    \includegraphics[width=6cm, angle=0]{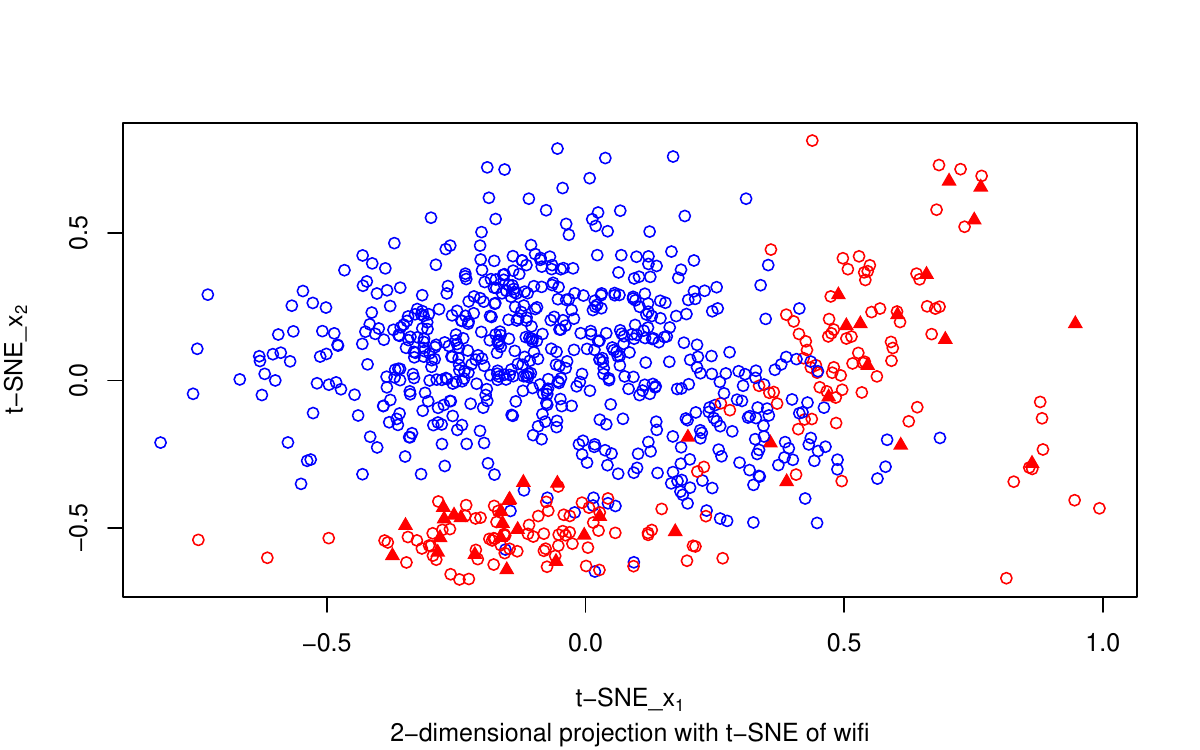}  
\includegraphics[width=6cm, angle=0]{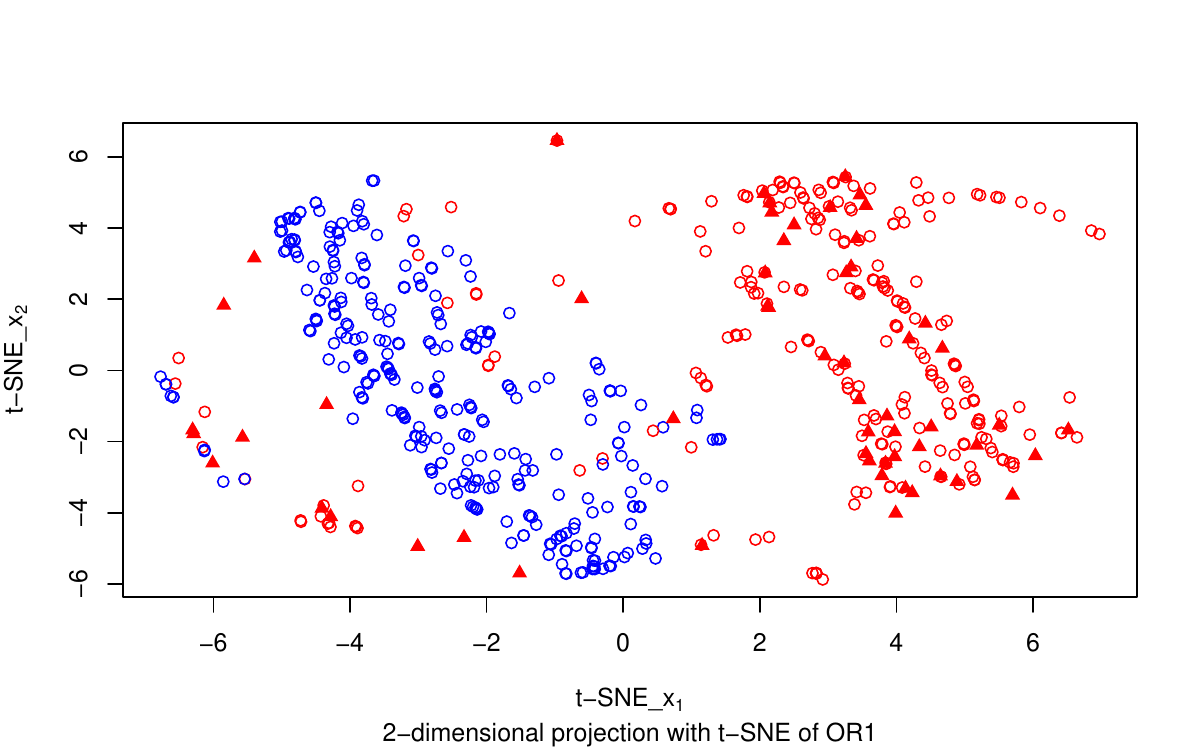}  
    \includegraphics[width=6cm, angle=0]{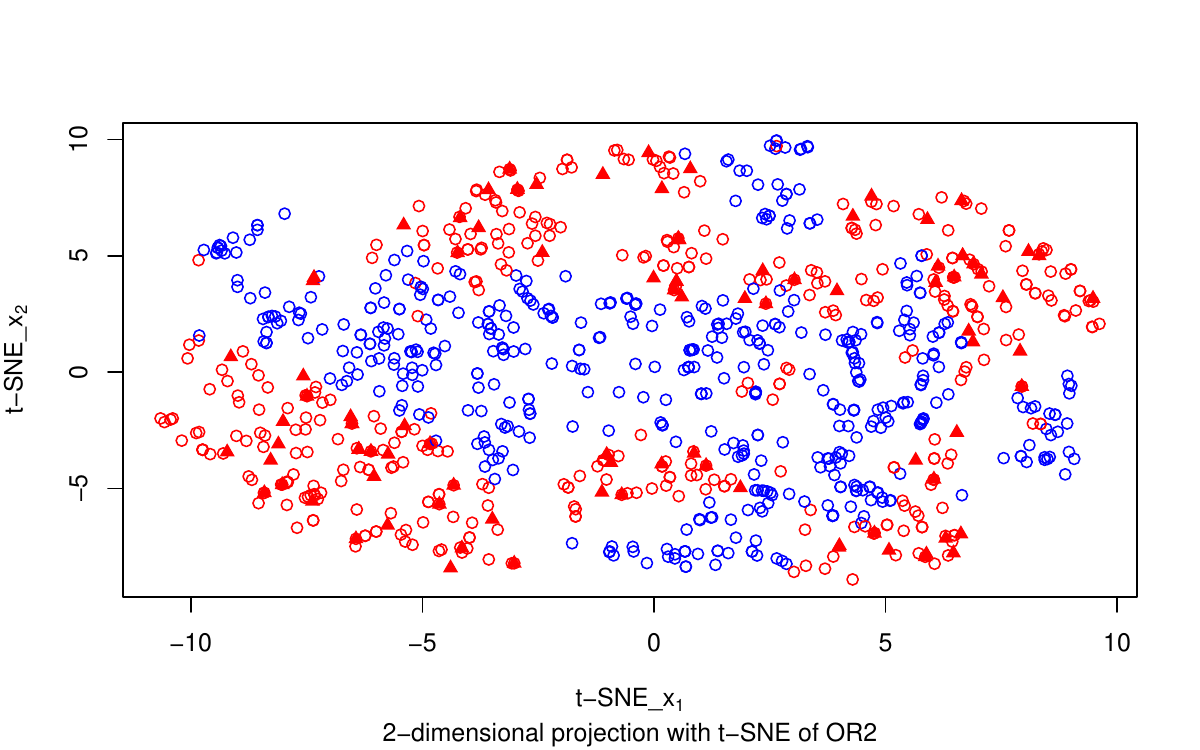}  
    \includegraphics[width=6cm, angle=0]{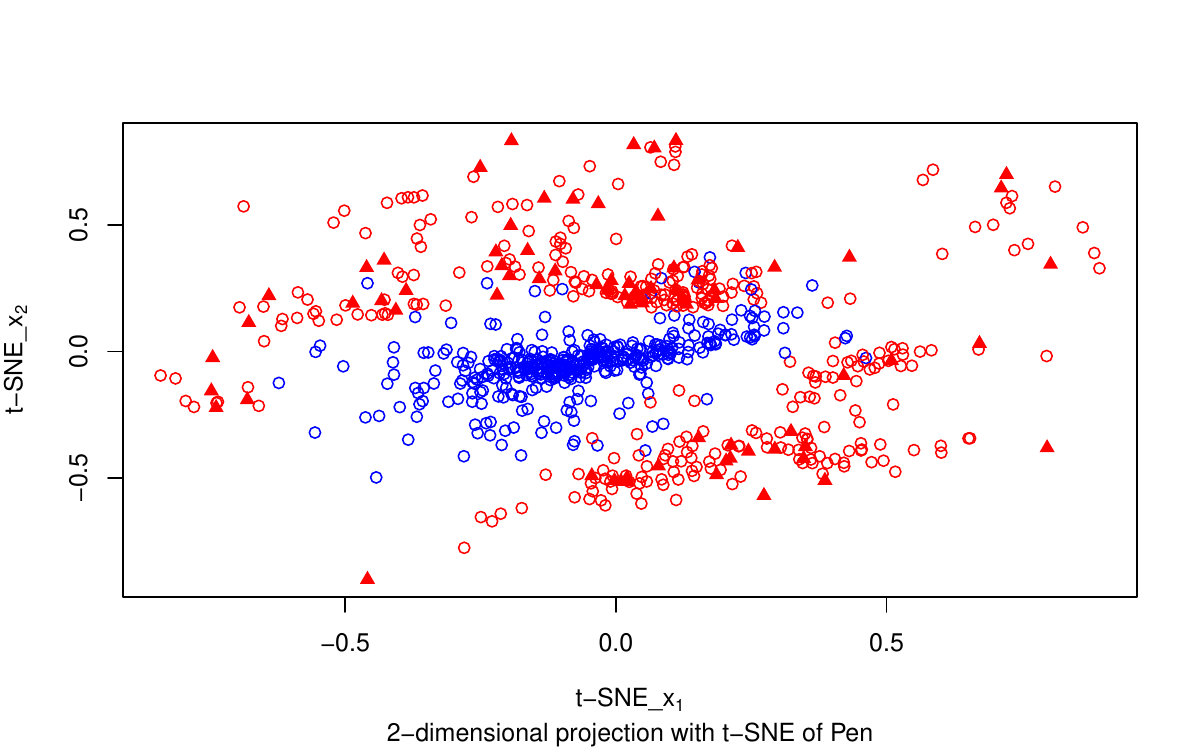}  

\caption{The t-SNE  plots of the best four datasets \textbf{wifi}, \textbf{OR1}, \textbf{OR2} and \textbf{Pen}; the perplexity for the training of t-SNE on these four datasets was set to 750, 40, 250, 750, respectively; label frequency $\gamma$=0.25; red: positive instances; blue: negative instances; triangle: labeled instances; circle: unlabeled instances.}          
\label{12}
\end{figure}  

The t-SNE plots of the best four datasets, i.e., \textbf{wifi}, \textbf{OR1}, \textbf{OR2}, and \textbf{Pen}, for PUAL compared with GLLC are shown in Fig.~\ref{12}. According to the four t-SNE plots, the following observations can be made:
\begin{enumerate}
    \item The trifurcate pattern of the  best four datasets in Fig.~\ref{12} is clear that the positive set is constituted by two subsets distributing on both sides of the negative set as discussed in the motivation of PUAL in Section \ref{motR}. 
    
   \item In the t-SNE plot of dataset \textbf{OR1}, there are many more instances in the right-hand labeled-positive subset than the instances in the left-hand labeled-positive set. This indicates that the trifurcate pattern does not have to be balanced for PUAL to outperform GLLC.

\end{enumerate}

\subsubsection{Pattern Analysis for the Real Datasets Preferring GLLC to PUAL }
\label{lllast}

\begin{figure}[htbp]    
\centering           
\includegraphics[width=6cm, angle=0]{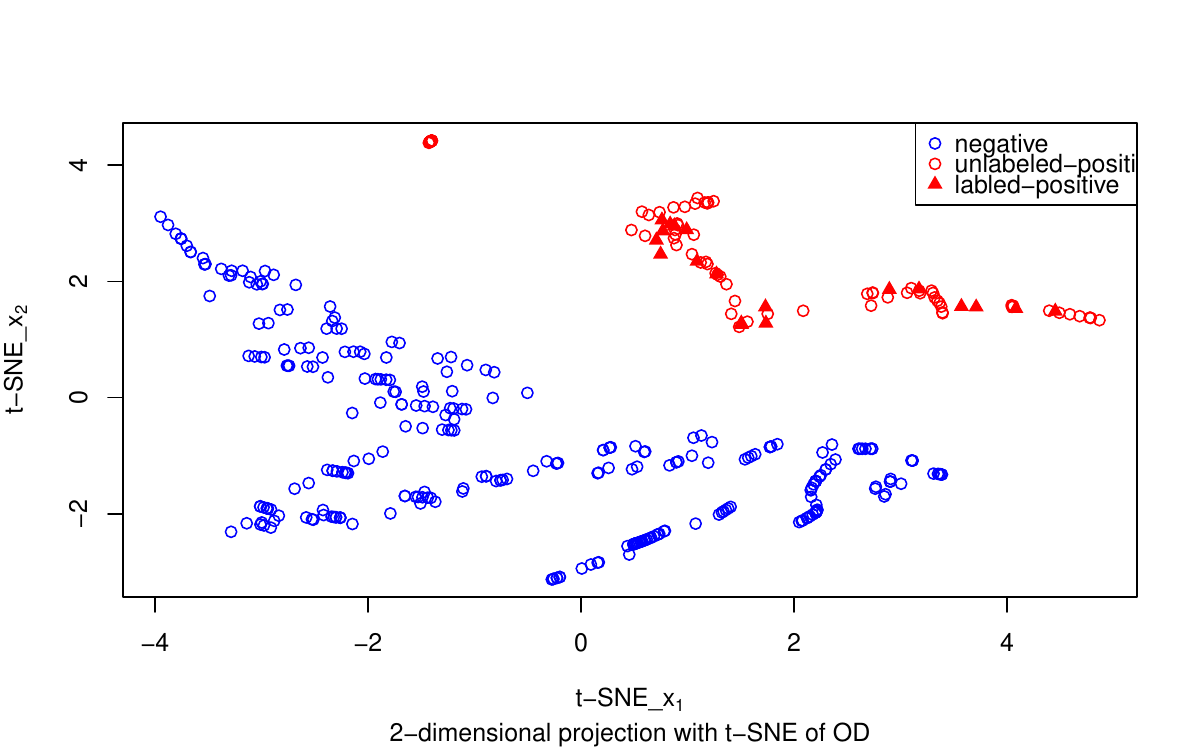}  
    \includegraphics[width=6cm, angle=0]{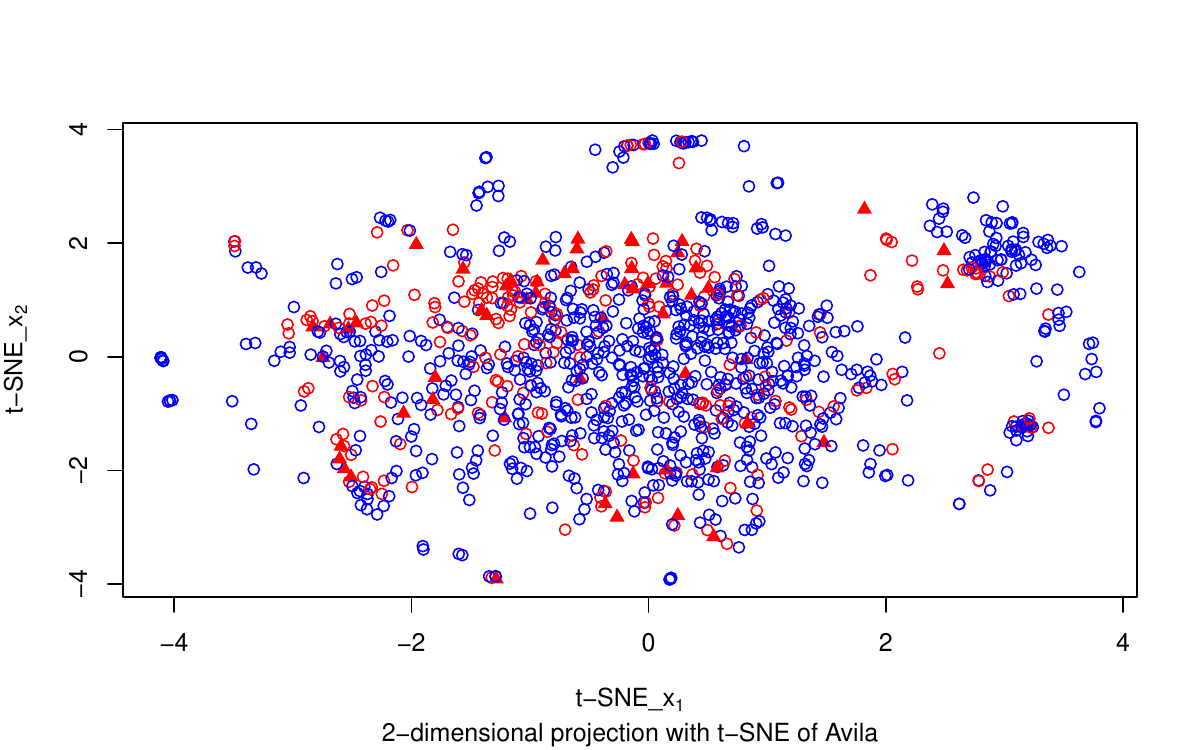} 
    \includegraphics[width=6cm, angle=0]{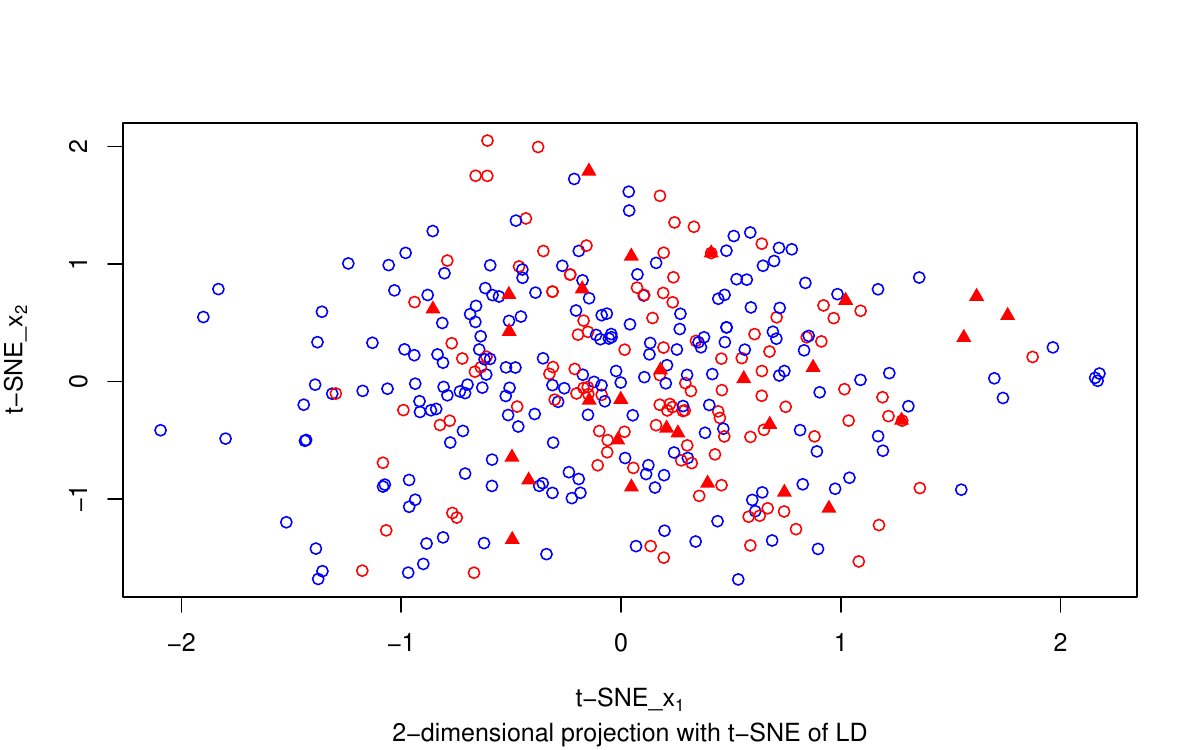}
    \includegraphics[width=6cm, angle=0]{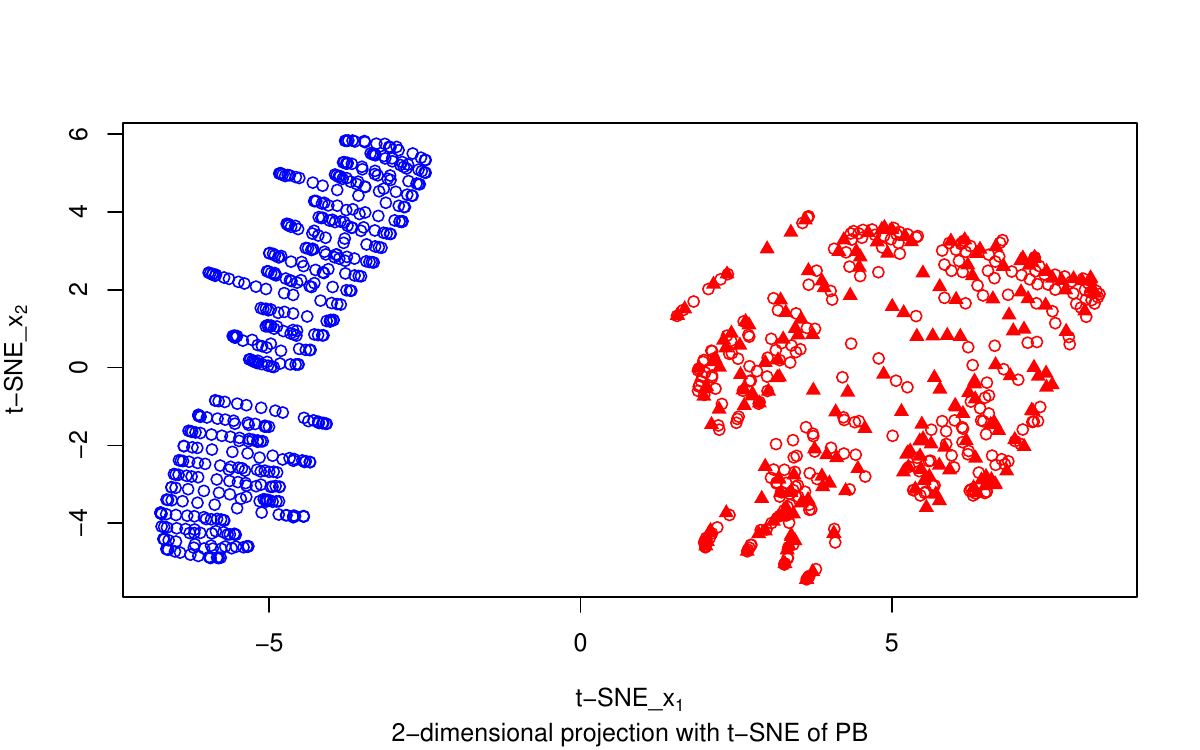}
    \includegraphics[width=6cm, angle=0]{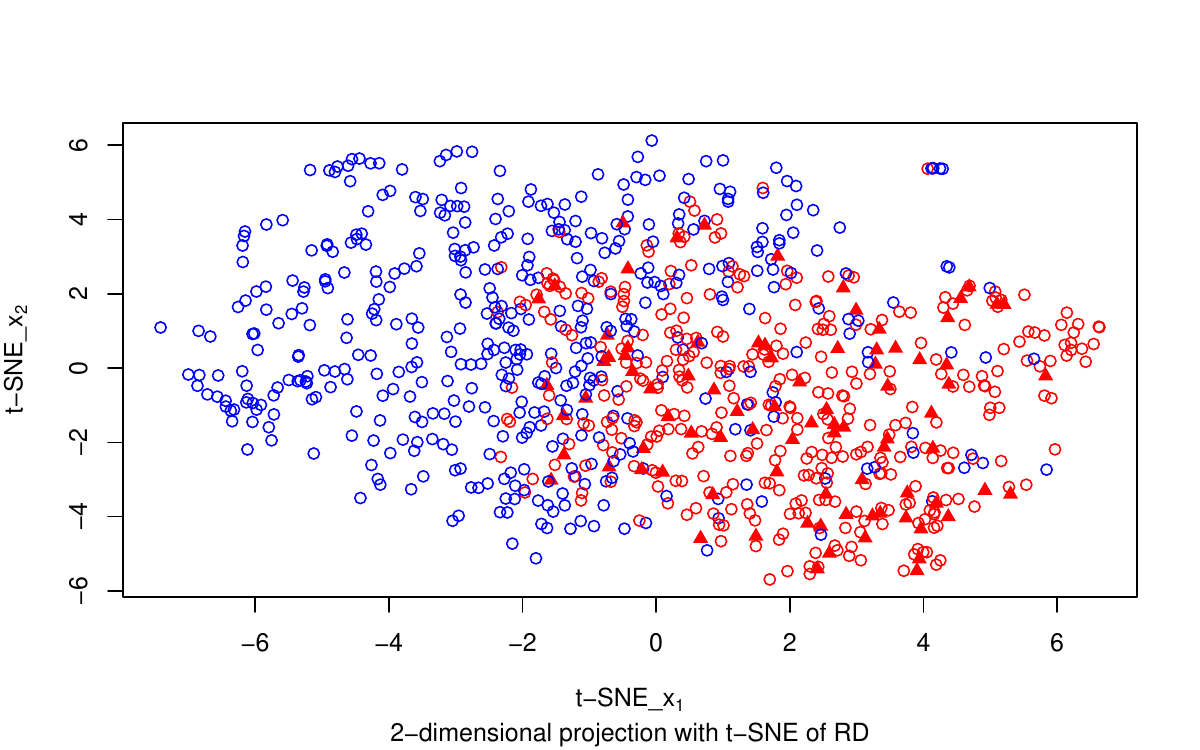}
\caption{ The t-SNE  plots of the worst five datasets \textbf{OD},  \textbf{Avila}, \textbf{LD},  \textbf{PB},  \textbf{RD}; cross entropy loss for training with perplexity =  200, 550, 250, 300, 300; label frequency $\gamma$=0.25, 0.25, 0.25, 0.5, 0.25. The rest of the caption is as in Fig.~\ref{12}. }          
\label{1653}
\end{figure}   

There are overall five datasets to be the worst four datasets for PUAL compared with GLLC under two cases of label frequency, i.e., $\gamma=0.5, 0.25$. The t-SNE plots for them are illustrated in Fig.~\ref{1653} and their patterns can be summarized as follows:
\begin{enumerate}
     \item According to the t-SNE plots of the two datasets  \textbf{LD} and \textbf{Avila},  the positive set and the negative set are mixed together, making the dataset challenging to  be separated. In this case,  the labeled-positive instances selected as support vectors by the hinge loss of PUAL are not sufficient to adequately represent the pattern of the positive set, while the squared loss of GLLC on the labeled-positive set, which selects all positive instances as the support vectors, can somewhat alleviate this issue. As a result, GLLC on these two datasets outperforms PUAL.

\item The t-SNE plots of the three datasets \textbf{OD},  \textbf{PB}, and \textbf{RD} represent the typical two-class patterns. In this type of datasets, the problem of GLLC mentioned in Section \ref{motR} does not exist. Therefore, the optimal combination(s) of the hyper-parameters for GLLC to outperform PUAL was (were) found in at least one case of label frequency $\gamma$.

\end{enumerate}

\subsubsection{Comparison between PUAL, uPU and nnPU}
 The performance of PUAL, GLLC, uPU and nnPU on the 16 real datasets are summarized in Table~\ref{resgl1}, from which we can obtain the following findings. First, there are in total 22 cases of the 32 cases  where PUAL outperforms uPU and nnPU. Secondly, uPU and nnPU sometimes have much larger standard deviations than PUAL since their algorithms based on Adam for the optimization of their non-convex objective functions  cannot always converge to the optimal solution, although  nnPU alleviates this issue to some extent. 

\renewcommand{\arraystretch}{0.6}
\begin{longtable}{l|cccccccc}
    \caption{The average F1-score (\%) of the classifiers. For each of the 16 original datasets, the average F1-scores and standard deviations in the two rows were obtained under label frequencies $\gamma=0.5$ (top) and $0.25$ (bottom), respectively. In each row, the result highlighted in blue indicates the statistically optimal choice among the four methods according to the t-tests with $p$-value lower than 0.05 (Sometimes an optimal choice was missed in a row since there was no such a method holding statistically significantly higher F1-score than other methods). } \label{resgl1}\\
\toprule
Dataset & {PUAL} &{GLLC} &{uPU} &{nnPU} \\
\midrule
\endfirsthead

\multicolumn{9}{c}%
{{\tablename\ \thetable{} -- continued from previous page}} \\
\toprule
Dataset & {PUAL} & {GLLC} & {uPU} & {nnPU} \\
\midrule
    \endhead

    \endfoot

    \multicolumn{1}{l|}{ENB} & 42.82 $\pm$ 4.76 & 42.69 $\pm$ 4.62 & 29.58     $\pm$ 22.14      & 30.20      $\pm$ 23.67 \\
          & 45.82 $\pm$ 7.50 & 44.16 $\pm$ 6.56 &   26.12    $\pm$ 30.53      &  26.88     $\pm$  31.28\\
          
    \midrule
    \multicolumn{1}{l|}{HD} & 82.72 $\pm$ 2.35 & 81.97 $\pm$ 5.45 & 71.38      $\pm$ 4.23      & 74.38      $\pm$ 2.19 \\
          & 81.92 $\pm$ 4.03 & \textcolor{blue}{84.46 $\pm$ 4.11} &   71.01    $\pm$  3.97     & 75.06     $\pm$ 2.40  \\
        
    \midrule
    \multicolumn{1}{l|}{Pen} &\textcolor{blue}{92.47 $\pm$ 8.13} & 88.92 $\pm$ 10.15 & 77.76      $\pm$ 31.00      & 87.50      $\pm$ 14.94 \\
          & \textcolor{blue}{91.73 $\pm$ 9.04}& 87.02 $\pm$ 11.38 & 72.55      $\pm$  31.03     & 84.06      $\pm$ 16.85 \\
         
    \midrule
    \multicolumn{1}{l|}{LD} & 44.24 $\pm$ 5.72 &\textcolor{blue}{50.79 $\pm$ 6.86}& 11.88      $\pm$ 25.75      &  31.54     $\pm$ 27.79\\
          & 36.85 $\pm$ 9.97 & \textcolor{blue}{40.05 $\pm$ 8.85} & 10.15      $\pm$  22.39     &   20.09    $\pm$ 26.27 \\
       \midrule  
    \multicolumn{1}{l|}{OR1} &\textcolor{blue}{90.05 $\pm$ 2.25}& 85.62 $\pm$ 3.77 & 16.64      $\pm$ 33.33      & 84.08  $\pm$ 6.88  \\
          &\textcolor{blue}{83.88 $\pm$ 5.78}& 72.95 $\pm$ 5.50 &  20.90     $\pm$  33.12     &  72.06     $\pm$ 6.99 \\
         \midrule
    \multicolumn{1}{l|}{RD} & 82.45 $\pm$ 2.18 & 83.01 $\pm$ 2.38 & 70.61      $\pm$ 12.89      & 71.29     $\pm$ 13.63 \\
          & 77.63 $\pm$ 3.77 & \textcolor{blue}{81.99 $\pm$ 2.84} &  72.92     $\pm$  14.53     & 73.12      $\pm$ 12.83 \\
     \midrule%

    \multicolumn{1}{l|}{Seeds} & 92.31 $\pm$ 4.86 & {94.63 $\pm$ 2.81} & 92.37      $\pm$  1.51    &  \textcolor{blue}{97.25    $\pm$ 3.65}  \\
          & 89.05 $\pm$ 5.53 & 91.18 $\pm$ 4.48 &  86.85    $\pm$  3.12     &  \textcolor{blue}{93.08    $\pm$ 3.89} \\
        \midrule
    \multicolumn{1}{l|}{wifi} &\textcolor{blue}{95.10 $\pm$ 1.96}& 90.43 $\pm$ 4.65 & 91.16     $\pm$ 4.29      & 92.17       $\pm$ 3.17 \\
          & \textcolor{blue}{96.69 $\pm$ 1.83}& 89.09 $\pm$ 4.77 &   87.69     $\pm$   2.88     & 89.27    $\pm$ 2.61 \\
    
    \midrule
    \multicolumn{1}{l|}{Avila} & 55.82 $\pm$ 2.90 & 59.56 $\pm$ 4.10 & 62.74      $\pm$  8.82     & 63.75      $\pm$ 9.07 \\
          & 50.05 $\pm$ 4.22 & 54.99 $\pm$ 6.17 &  61.30    $\pm$ 9.23      &  61.00     $\pm$ 9.04 \\
   
    \midrule
    \multicolumn{1}{l|}{OD} & 89.00 $\pm$ 8.44 & 100.00 $\pm$ 0.00 &  80.00     $\pm$   42.16    &   100.00    $\pm$ 0.00 \\
          & 95.69 $\pm$ 6.74 & 100.00 $\pm$ 0.00 & 80.00      $\pm$   42.16    &  100.00     $\pm$  0.00\\
  
    \midrule
    \multicolumn{1}{l|}{OR2} &\textcolor{blue}{88.93 $\pm$ 1.22}& 86.49 $\pm$ 1.38 &  76.92      $\pm$ 4.90      &  81.60    $\pm$ 4.23 \\
          & \textcolor{blue}{85.50 $\pm$ 3.42}& 77.10 $\pm$ 5.65 & 74.41      $\pm$  5.45     &  77.28     $\pm$ 3.76 \\
     
    \midrule
    \multicolumn{1}{l|}{PB} & 95.90 $\pm$ 1.12 & \textcolor{blue}{100.00 $\pm$ 0.00} & 69.77      $\pm$  2.62     & 67.19      $\pm$ 3.17 \\
          & 97.86 $\pm$ 0.67 & \textcolor{blue}{100.00 $\pm$ 0.00} &  68.75     $\pm$ 2.63      &   66.63    $\pm$ 4.09 \\
    
    \midrule
    
    \multicolumn{1}{l|}{Acc} & 65.02 $\pm$ 4.79 & \textcolor{blue}{68.10 $\pm$ 2.18} & 20.05     $\pm$ 27.57       &  20.46    $\pm$ 28.56 \\
          & 66.36 $\pm$ 4.42 & 64.08 $\pm$ 6.31 &21.95       $\pm$ 29.61      &  23.43     $\pm$ 31.40 \\
     
    \midrule
    \multicolumn{1}{l|}{Ecoli} & 90.80 $\pm$ 2.59 & 90.15 $\pm$ 6.28 &84.41       $\pm$ 6.13      & 85.92      $\pm$  6.69 \\
          & 88.04 $\pm$ 4.44 & 88.03 $\pm$ 3.96 &   84.92    $\pm$  6.82     &  86.05     $\pm$  6.55 \\
    
    \midrule
    
    \multicolumn{1}{l|}{SSMCR} & 87.63 $\pm$ 1.29 & 87.35 $\pm$ 2.14 & 85.71      $\pm$ 1.98      & 87.42      $\pm$ 1.37 \\
          & 87.63 $\pm$ 1.29 & 87.35 $\pm$ 2.14 &  84.97     $\pm$ 2.03       &   86.79    $\pm$ 1.50 \\
    
    \midrule
    \multicolumn{1}{l|}{UMD} & 100.00 $\pm$ 0.00 & 99.80 $\pm$ 0.62 & 100.00     $\pm$ 0.00      & 100.00      $\pm$ 0.00  \\
          & 99.58 $\pm$ 0.88 & 98.41 $\pm$ 1.80 & 100.00      $\pm$ 0.00     & 100.00      $\pm$ 0.00 \\
       \bottomrule
\end{longtable}
\renewcommand{\arraystretch}{1}

\section{Conclusion}
In this paper, we propose PUAL for better classification on trifurcate PU datasets, where the positive set is constituted by two subsets distributing on both sides of the negative set. The key novelty of PUAL is an asymmetric structure composed of hinge loss and squared loss for training the PU classifier.  Experimental results demonstrate the superiority of PUAL in performing PU classification on trifurcate data. As an SVM-based method, PUAL is still negatively affected by the irrelevant features in the datasets, partly due to the L2 regularization term of model parameters, which cannot compress the coefficients of the irrelevant features to be exactly zero thoroughly~\cite{nguyen2011general}. To address this issue, a future work is to explore the  L1 regularization term of the model parameters for PUAL.

\bibliography{ref}

\end{document}